\documentclass[letterpaper]{article} 
\usepackage{aaai24}  
\usepackage{times}  
\usepackage{helvet}  
\usepackage{courier}  
\usepackage[hyphens]{url}  
\usepackage{graphicx} 
\usepackage{natbib}  
\usepackage{caption} 
\usepackage{multirow}
\usepackage[table,xcdraw]{xcolor}
\usepackage{booktabs}
\usepackage{appendix}
\usepackage{algorithm}
\usepackage{algorithmic}
\usepackage{amssymb}
\usepackage{amsmath}
\usepackage{diagbox}

\definecolor{boxhead}{rgb}{0.70196078, 0.80392157, 0.89019608}
\definecolor{boxbody}{rgb}{0.9215, 0.9215, 0.9215}
\definecolor{boxus}{rgb}{0.9961, 0.8510, 0.6510}

\usepackage{xcolor}

\newcommand{\etal}{{\textit{et al.}}}

\usepackage{newfloat}
\usepackage{listings}
\DeclareCaptionStyle{ruled}{labelfont=normalfont,labelsep=colon,strut=off} 
\floatstyle{ruled}
\newfloat{listing}{tb}{lst}{}
\floatname{listing}{Listing}
\makeatletter
\renewcommand{\maketag@@@}[1]{\hbox{\m@th\normalsize\normalfont#1}}%
\makeatother

\title{CMG-Net: Robust Normal Estimation for Point Clouds via \\Chamfer Normal Distance and Multi-scale Geometry}

\author {
    Yingrui Wu\textsuperscript{\rm 1,\rm 2}\equalcontrib,
    Mingyang Zhao\textsuperscript{\rm 3}\equalcontrib,
    Keqiang Li\textsuperscript{\rm 4},
    Weize Quan\textsuperscript{\rm 1,\rm 2},
    Tianqi Yu\textsuperscript{\rm 5},
    Jianfeng Yang\textsuperscript{\rm 5},
    Xiaohong Jia\textsuperscript{\rm 6,\rm 2},
    Dong-Ming Yan\textsuperscript{\rm 1,\rm 2} \footnote{Corresponding author.}
}
\affiliations {
    \textsuperscript{\rm 1}MAIS, Institute of Automation, Chinese Academy of Sciences,
    \textsuperscript{\rm 2}University of Chinese Academy of Sciences \\
    \textsuperscript{\rm 3}Hong Kong Institute of Science \& Innovation, Chinese Academy of Sciences,
    \textsuperscript{\rm 4}SenseTime Research\\
    \textsuperscript{\rm 5}School of Electronic and Information Engineering, Soochow University,
    \textsuperscript{\rm 6}AMSS, Chinese Academy of Sciences\\
    wuyingrui2023@ia.ac.cn, \{migyangz, likeq98, qweizework, yandongming\}@gmail.com \\\{tqyu, jfyang\}@suda.edu.cn, xhjia@amss.ac.cn
}

\usepackage{bibentry}

\begin{document}

\maketitle

\begin{abstract}
This work presents an accurate and robust method for estimating normals from point clouds. In contrast to predecessor approaches that minimize the deviations between the annotated and the predicted normals directly, leading to direction inconsistency, we first propose a new metric termed \emph{Chamfer Normal Distance} to address this issue. This not only mitigates the challenge but also facilitates network training and substantially enhances the network robustness against noise. Subsequently, we devise an innovative architecture that encompasses \emph{Multi-scale Local Feature Aggregation} and \emph{Hierarchical Geometric Information Fusion}. This design empowers the network to capture intricate geometric details more effectively and alleviate the ambiguity in scale selection. Extensive experiments demonstrate that our method achieves the state-of-the-art performance on both synthetic and real-world datasets, particularly in scenarios contaminated by noise. Our implementation is available at \url{https://github.com/YingruiWoo/CMG-Net_Pytorch}.
\end{abstract}

\section{Introduction}\label{sec:intro}
Normal estimation is a fundamentally important task in the field of point cloud analysis, which enjoys a wide variety of applications in 3D vision and robotics, such as surface reconstruction~\cite{fleishman2005robust,kazhdan2006poisson}, denoising~\cite{lu2020low} and semantic segmentation~\cite{grilli2017review,che2018multi}. In recent years, many powerful methods have been developed to enhance the performance of normal estimation. However, these approaches involving both traditional and learning-based ones often suffer from \emph{heavy noise} and struggle to attain high-quality results for point clouds with \emph{complex geometries}.

\begin{figure}[t]
\centering
\includegraphics[scale=0.33]{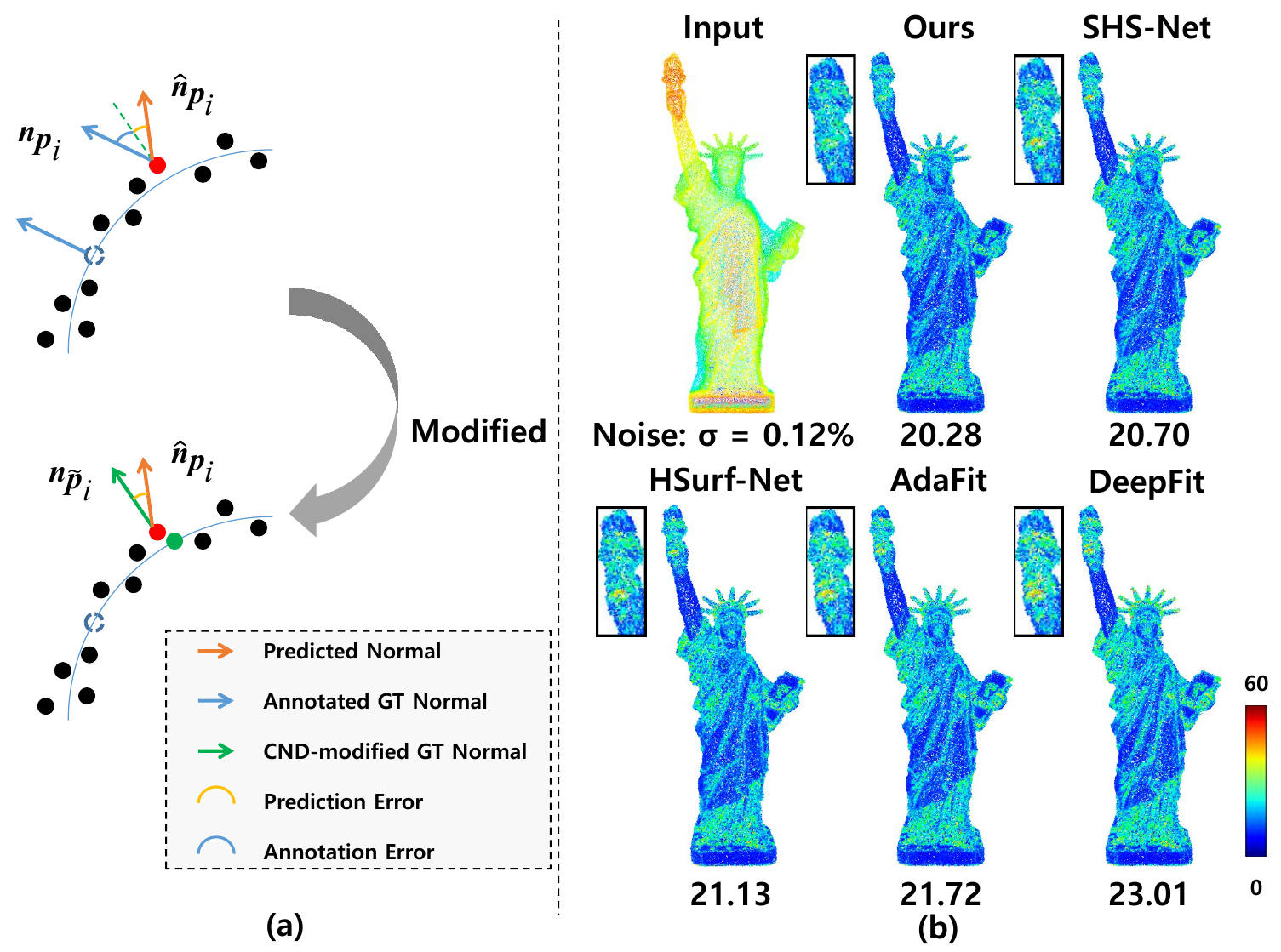}
\caption{(a) Comparison between the annotated and the proposed CND-modified normals, where the latter is more consistent with the underlying surface geometry. (b) Our method outperforms competitors with higher robustness to noise and intricate shape details (indicated by the heat map).}
\label{fig: methods_comparison}
\end{figure}

Traditional methods~\cite{hoppe1992surface,levin1998approximation,cazals2005estimating} typically encompass fitting local planes or polynomial surfaces and inferring normal vectors from the fitted outcomes. Although straightforward, these approaches are vulnerable to noise and encounter challenges when attempting to generalize to complex shapes. Furthermore, their performance hinges significantly on the meticulous tuning of parameters.

In comparison with traditional approaches, learning-based proposals~\cite{guerrero2018pcpnet,ben2019nesti,hashimoto2019normal,zhou2020geometry,wang2020neighbourhood,lenssen2020deep,ben2020deepfit,cao2021latent,zhu2021adafit,zhou2022fast,zhang2022geometry,li2022graphfit,li2022hsurf,du2023rethinking,li2023shs} have better generalization and less dependency on parameter tuning. There are two types of learning-based normal estimators comprising \emph{deep surface fitting} and \emph{regression}. The former predicts point-wise weights of the input point cloud patch and derives a polynomial surface by \emph{weighted least-squares} (WLS)  fitting. However, due to the fixed order of polynomial functions, deep surface fitting usually grapples with overfitting or underfitting when dealing with various surfaces. On the contrary, regression-based methods adopt \emph{Multi-Layer Perception} (MLP) to extract features of the input patch and directly regress normal vector from these features. Benefiting from their strong feature extraction capability, recent regression-based methods have advanced normal estimation for clean point clouds. However, they have not made substantial headway in improving normal estimation on point clouds that are affected by noise.

To address noisy normal estimation issues, in this paper, we first analyze the normal estimation deviation produced by noisy point clouds, and then point out the \emph{inconsistency} between the annotated (ground-truth) normal and the input patch, as illustrated in Fig.~\ref{fig: methods_comparison}(a). We find that this direction inconsistency indeed significantly affects the network training and the output evaluation, also degrading the downstream tasks such as surface reconstruction. The reason is that when point coordinates change significantly due to noisy influence, their neighborhood geometry and normals change accordingly, while the annotated normals are fixed. To deal with this problem, instead of using the conventional \emph{Root of Mean Squared Error} (RMSE), we propose a more reasonable metric for normal estimation termed \emph{Chamfer Normal Distance} (CND), which replaces the original annotated normal vector with the normal of the closest point locating on the potential clean point cloud. Moreover, we minimize the loss function modified by CND to reduce the disturbance of inconsistency deviations during training. We show that our newly defined loss function achieves much higher normal estimation accuracy than competitors on a set of benchmark datasets.

Furthermore, we develop a novel network framework named CMG-Net, integrating the CND-modified loss with multi-scale geometric structures, for more stable and robust normal estimation. Unlike previous approaches that merely capture a single scale of local or global features, CMG-Net performs a multi-scale feature extraction followed by integration through an attention layer. This greatly facilitates the network capability to capture intricate geometric details and address the ambiguity of the optimal scale selection. Moreover, we further combine the local and global features from various scales together in the hierarchical architecture to increase the multi-scale information for network inference.

We conduct extensive experiments to validate the developed method and compare it with the \emph{state-of-the-art} (SOTA) approaches on various benckmark datasets including \emph{PCPNet}~\cite{guerrero2018pcpnet} and the \emph{indoor SceneNN} dataset~\cite{hua-pointwise-cvpr18}. Results demonstrate that our method outperforms baselines by a large margin, especially on point clouds with noise, and those with intricate geometric details and various distribution density. To summarize, our main technical contributions are threefold as follows:

\begin{itemize}
    \item We propose a new method that integrates the CND metric for robust normal estimation, which solves the direction inconsistency problem effectively and significantly boosts network training and inference.
    \item  We design a novel network that incorporates multi-scale feature extraction along with hierarchical inference combined with intricate geometry information fusion, which is capable of capturing intricate geometric details and addressing the challenge of scale selection ambiguity.
    \item We perform comprehensive experiments to demonstrate the enhancements brought by our proposed method, thereby pushing the boundaries of SOTA performance, especially on noisy normal estimation scenarios.
\end{itemize}

\section{Related Work}

\subsection{Traditional Methods}
{Principal Component Analysis} (PCA)~\cite{hoppe1992surface} stands as the most widely adopted point cloud normal estimation method, which fits a plane to the input surface patch. Subsequent variants involving {Moving Least Squares} (MLS)~\cite{levin1998approximation}, {truncated Taylor expansion fitting} (n-jet)~\cite{cazals2005estimating}, local spherical surface fitting~\cite{guennebaud2007algebraic} and multi-scale kernel~\cite{aroudj2017visibility} are proposed to reduce the noisy influence through selecting larger patches and employing more intricate energy functions. Nevertheless, these approaches typically tend to oversmooth sharp features and geometric details. To circumvent these issues, Voronoi diagram~\cite{amenta1998surface,alliez2007voronoi,merigot2010voronoi}, Hough transform~\cite{boulch2012fast}, and plane voting~\cite{zhang2018multi} are deployed in normal estimation. However, these techniques depend on manual parameter tuning heavily, which hinders their practical applications.

\subsection{Learning-based Methods}
With the powerful development of neural network, learning-based normal estimation achieves better performance and less dependence of parameter tuning than traditional approaches. They can be generally divided into two categories: \emph{deep Surface fitting} and \emph{regression-based} approaches.

\subsubsection{Deep surface fitting methods.}
These methods typically employ a deep neural network to predict point-wise weights and then fit a polynomial surface to input patches using WLS such as IterNet~\cite{lenssen2020deep} and DeepFit~\cite{ben2020deepfit}. Analogously, Zhang~\etal~\shortcite{zhang2022geometry} adopted the predicted weights as the guiding geometric information. AdaFit~\cite{zhu2021adafit} proposed a novel layer to aggregate features from multiple global scales and then predicted point-wise offset to improve the normal estimation accuracy. To learn richer geometric features, GraphFit~\cite{li2022graphfit} combined graph convolutional layers with adaptive modules, while Du~\etal~\shortcite{du2023rethinking} analyzed the  approximation error of these methods and suggested two fundamental design principles to further improve the estimation accuracy. However, due to the constant order of the objective polynomial functions, deep surface fitting methods typically suffer from overfitting and underfitting.

\subsubsection{Regression-based methods.}
This type casts the normal estimation problem as a regression process and predicts the point cloud normals via the network straightforward. For instance, HoughCNN~\cite{boulch2016deep} transformed point clouds into a Hough space and then utilized {Convolutional Neural Networks} (CNN) to directly infer normal vectors, whereas Lu~\etal~\shortcite{lu2020deep} projected point clouds into a height map by computing distances between scatter points and the fitted plane. However, these approaches sacrifice the 3D geometry unavoidably when executing in 2D spaces. PCPNet~\cite{guerrero2018pcpnet} directly adopted the unstructured point clouds as input and then used the PointNet~\cite{qi2017pointnet} to capture multi-scale features instead. Hashimoto~\etal~\shortcite{hashimoto2019normal}  combined PointNet with 3D-CNN to extract local and spatial features, and NestiNet~\cite{ben2019nesti} employed mixture-of-experts framework to determine the optimal normal estimation scale. To provide more information of the input patch, Refine-Net~\cite{zhou2022refine} additionally calculated the initial normals and the height map. Recent work involve HSurf-Net~\cite{li2022hsurf} and SHS-Net~\cite{li2023shs} first transformed point clouds into a hyper space through local and global feature extractions and then performed plane fitting in the constructed space. 
NeAF~\cite{li2023neaf} inferred an angle field around the ground truth normal to make it learn more information of the input patch. Benefiting from the strong feature extraction abilities of the network architectures, recent regression-induced approaches demonstrate promising results on clean point clouds. However, they have yet made significant progress in normal estimation on noisy point clouds, which are often emerged in practical scenarios.

Aiming at improving the robustness to noise, we identify a crucial inconsistency between the annotated normal and the neighborhood geometry of the noisy point and introduce CND to address this problem. Besides, compared with the recent regression methods, we propose a network that combines various geometric information extraction with a hierarchical architecture to make the complex information capture more effectively.

\section{Rethinking Noisy Normal Estimation}

\subsection{Direction Inconsistency}
\begin{figure}[t]
\centering
\includegraphics[scale=0.397]{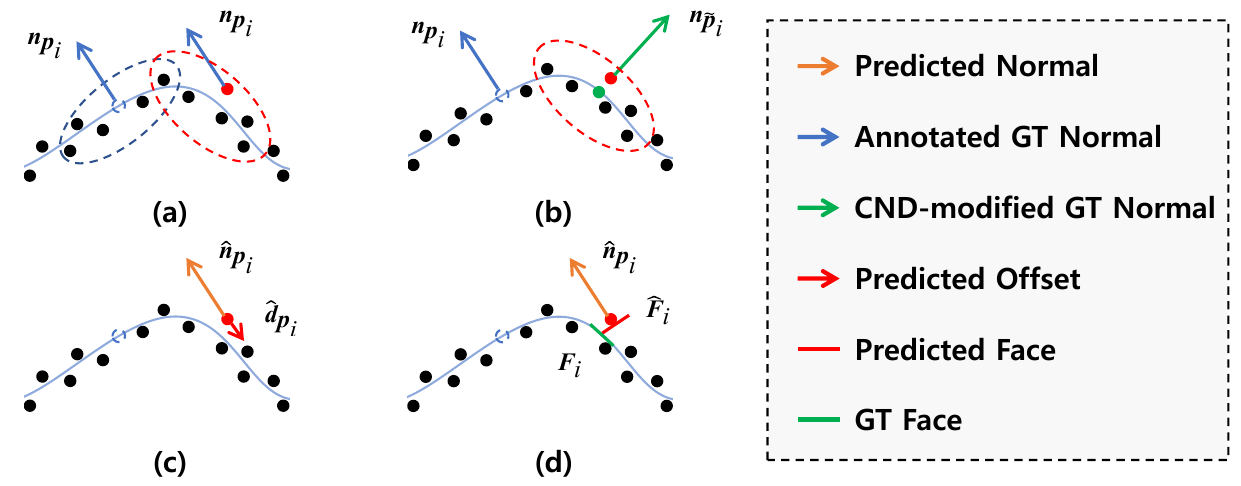}
\caption{(a) The annotated normal $\boldsymbol{n}_{\boldsymbol{p}_i}$ of noisy point $\boldsymbol{p}_i$ determined before noisy disturbance indeed is inconsistent with the input patch (dashed red ellipse). (b) The direction of the normal $\boldsymbol{n}_{\tilde{p}_i}$ of the nearest clean point $\tilde{\boldsymbol{p}}_i$ is more consistent with the input patch. (c) The predicted offset $\boldsymbol{\hat{d}}_{\boldsymbol{p}_i}$ cannot drag $\boldsymbol{p}_i$ to the noise-free underlying surface. (d) This inconsistency also arises for surface reconstruction assignments. }
\label{fig: comparison_between_the_ground-truth_normals}
\end{figure}

Previous learning-based approaches directly minimize the deviations between the predicted normals and the annotated ones for training and evaluation. This is reasonable for noise-free scenarios, however, for the noisy point clouds, due to the noise-caused relative coordinate changes, the annotated normals indeed are inconsistent with the neighborhood geometry of the query points. As presented in Fig.~\ref{fig: comparison_between_the_ground-truth_normals}(a), given a set of noisy point clouds $\mathcal{P}$, suppose the ground truth position locating on the surface of the noisy point $\boldsymbol{p}_i$ is $\tilde{\boldsymbol{p}}_i$. The annotated normal of $\boldsymbol{p}_i$ is $\boldsymbol{{n}}_{\boldsymbol{p}_i}\in \mathbb{R}^3$, which is the same as the one of the point before adding noise, and the normal of $\tilde{\boldsymbol{p}}_i$ is $\boldsymbol{n}_{\tilde{\boldsymbol{p}}_i}\in \mathbb{R}^3$. If we optimize the typically defined normal estimation loss $\|\boldsymbol{n}_{\boldsymbol{p}_i}-\boldsymbol{\hat{n}}_{\boldsymbol{p}_i}\|_2^2$ as predecessors, where  $\hat{\boldsymbol{n}}_{\boldsymbol{p}_i}$ is the predicted normal, this will unavoidably lead to inconsistency between the annotated normal $\boldsymbol{{n}}_{\boldsymbol{p}_i}$ and the input patch $\boldsymbol{P}_i$. What's worse, this inconsistency greatly decreases the quality of the training data and thus lowers down the estimation ability of the network on noisy point clouds.

Moreover, this inconsistency also degrades downstream tasks such as denoising and 3D reconstruction. For instance, Fig.~\ref{fig: comparison_between_the_ground-truth_normals}(c) shows
the denosing principle for point clouds. If we utilize the predicted normal vector $\hat{\boldsymbol{n}}_{\boldsymbol{p}_i}$, which closely resembles the annotated normal vector $\boldsymbol{n}_{\boldsymbol{p}_i}$ (indicating a highly accurate estimation), then the introduced offset $\boldsymbol{\hat{d}}_{\boldsymbol{p}_i}$ will not align or bring $\boldsymbol{p}_i$ closer to the noise-free underlying surface. Anonymously, in the context of reconstruction tasks, as shown in Fig.~\ref{fig: comparison_between_the_ground-truth_normals}(d), the regenerated mesh face $\boldsymbol{\hat{F}}_i$ in relation to the normal vector $\boldsymbol{\hat{n}}_{\boldsymbol{p}_i}$ significantly deviates from the authentic mesh fact $\boldsymbol{F}_i$.

\subsection{Scale Ambiguity}
Another challenge in current normal estimation approaches is the ambiguity regarding the optimal scale in both local and global feature extraction. Concerning local structures, using large scales typically improves robustness against noise but can lead to oversmoothing of shape details and sharp features. Conversely, small scales can preserve geometric details but are relatively sensitive to noise. When it comes to global features, large scales include more structure information from the underlying surface but may also incorporate irrelevant points, thus degrading the geometry information of the input patch. On the other hand, small scales reduce irrelevant points but are less robust to noise. Previous works have struggled to effectively extract and combine multi-scale local and global features,
making them highly dependent on scale selection and resulting in unsatisfactory performance on both noisy point clouds and complex shape details.

\section{Proposed Method}

To solve the aforementioned issues, we propose a novel normal estimation approach that is robust against noise and less sensitive to scale selection. Concrete technical contributions are presented in the following.

\subsection{Chamfer Normal Distance}

\begin{figure*}[htbp]
\centering
\includegraphics[scale=0.44]{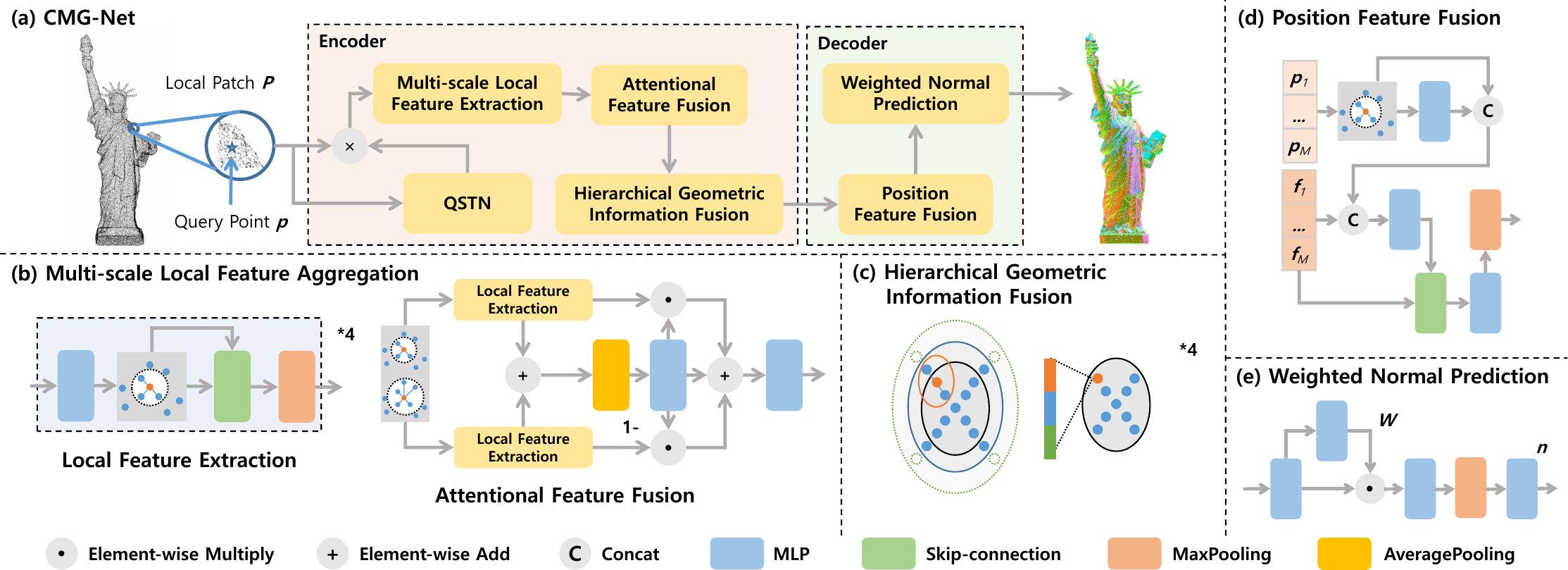}
\caption{Architecture of the proposed method. (a) Overall structure of CMG-Net. (b) Multi-scale Local Feature Aggregation. (c) Hierarchical Geometric Information Fusion. (d) Position Feature Fusion. (e) Weighted Normal Prediction.}
\label{fig: architecture_of_CMG-Net}
\end{figure*}

To bridge the direction inconsistency between the annotated normal and the predicted one of the input patch, instead of using the conventional metric $\|\boldsymbol{n}_{\boldsymbol{p}_i}-\hat{\boldsymbol{n}}_{\boldsymbol{p}_i}\|_2^2$, inspired from the \emph{Chamfer Distance} (CD)
\begin{equation}
\fontsize{6.5pt}{\baselineskip}\selectfont
\mathrm{C}(\mathcal{P},\hat{\mathcal{P}})
\!=\!\frac{1}{N_1}\sum_{\boldsymbol{p}_i\in\mathcal{P}}\operatorname*{min}_{\hat{\boldsymbol{p}}_j\in\hat{\mathcal{P}}}(\|\boldsymbol{p}_i\!-\!\hat{\boldsymbol{p}}_j\|_2^{2})\!+\!\frac{1}{N_2}\sum_{\boldsymbol{\hat{p}}_j\in\mathcal{\hat{P}}}\min_{\boldsymbol{p}_i\in\mathcal{P}}(\|\boldsymbol{p}_i\!-\!\hat{\boldsymbol{p}}_j\|_2^{2}),
\end{equation}
where $N_1$ and $N_2$ represent the cardinalities of the point cloud $\mathcal{P}$ and $\hat{\mathcal{P}}$, we formulate the \emph{Chamfer Normal Distance} (CND) as
\begin{equation}
\mathrm{CND}(\mathcal{P},{\tilde{\mathcal{P}}})=\sqrt{\frac{1}{N}\sum_{i=1}^{N}\mathrm{arccos}^2<\boldsymbol{n}_{\boldsymbol{\tilde{p}}_i}, \boldsymbol{\hat{n}}_{\boldsymbol{p}_i}>},
\end{equation}
where $<\cdot,\cdot>$ represents the inner product of two vectors and $\boldsymbol{\tilde{p}}_i$ is the closest point of $\boldsymbol{p}_i$ in the noise-free point cloud $\tilde{\mathcal{P}}$. In contrast to previous approaches that relied on annotated normal correspondence, our proposed CND manner assures consistency with the underlying geometric structure of the input patch (Fig.~\ref{fig: comparison_between_the_ground-truth_normals}(b)). The CND metric not only faithfully captures the prediction errors in noisy point clouds, but also  eliminates the direction inconsistency during network training, thus substantially improving the network robustness and facilitating the subsequent assignments.

\subsection{CMG-Net}
To capture more fruitful multi-scale structure information and solve the scale ambiguity issue simultaneously, we develop a network combining various geometric information extraction with a hierarchical architecture termed CMG-Net. Given a patch $\boldsymbol{P}=\{\boldsymbol{p}_i\in\mathbb{R}^3\}_{i=1}^N$ centralized at a query point $\boldsymbol{p}$, as shown in Fig.~\ref{fig: architecture_of_CMG-Net}(a), CMG-Net first normalizes the input points and rotates $\boldsymbol{P}$ by PCA and QSTN~\cite{qi2017pointnet, du2023rethinking} to initialize the normal vectors. Then, we group the local features by \emph{k-nearest neighbors} ($k$-NN) with different scales and aggregate them together. Besides, we design a hierarchical structure with intricate geometry information fusion, followed by the decoding of the embedding features. Our loss function modified by CND enables the network jumping out of the annotation inconsistency.

\subsubsection{Multi-scale Local Feature Aggregation.}

\begin{table*}[htbp]
\centering
\caption{Quantitative comparisons in terms of RMSE and CND on the PCPNet dataset. \textbf{Bold} values indicate the best estimator.}
\scalebox{0.71}[0.71]{
\begin{tabular}{l|cccc|cc|c|cccc|cc|c}
\hline
{\multirow{3}{*}{Method}}&\multicolumn{7}{c|}{{{RMSE}}}& \multicolumn{7}{c}{{CND}}  \\ \cline{2-15}
& \multicolumn{4}{c|}{{Noise ($\sigma$)}}& \multicolumn{2}{c|}{{Density}}& \multirow{2}{*}{{Ave.}}       & \multicolumn{4}{c|}{{Noise ($\sigma$)}}& \multicolumn{2}{c|}{{Density}} & \multirow{2}{*}{{Ave.}}  \\
&None          & 0.12\%        & 0.6\%          & \multicolumn{1}{c|}{1.2\%}          & Stripes       & \multicolumn{1}{c|}{Gradient}      & & None          & 0.12\%        & 0.6\%          & \multicolumn{1}{c|}{1.2\%}          & Stripes    & \multicolumn{1}{c|}{Gradient}      & \\ \hline
\hline
PCA~\cite{hoppe1992surface}                                             & 12.28         & 12.86         & 18.40          & \multicolumn{1}{c|}{27.61}          & 13.63         & \multicolumn{1}{c|}{12.79}         & 16.26                    & 12.28         & 12.78         & 16.41          & \multicolumn{1}{c|}{24.46}          & 13.63         & \multicolumn{1}{c|}{12.79}         & 15.39                    \\
n-jet~\cite{cazals2005estimating}                                           & 12.32         & 12.82         & 18.34          & \multicolumn{1}{c|}{27.77}          & 13.36         & \multicolumn{1}{c|}{13.09}         & 16.29                    & 12.32         & 12.77         & 16.36          & \multicolumn{1}{c|}{24.67}          & 13.36         & \multicolumn{1}{c|}{13.09}         & 15.43                    \\
PCPNet~\cite{guerrero2018pcpnet}                                          & 9.62          & 11.36         & 18.89          & \multicolumn{1}{c|}{23.32}          & 11.15         & \multicolumn{1}{c|}{11.69}         & 14.34                    & 9.62          & 11.23         & 17.28          & \multicolumn{1}{c|}{20.16}          & 11.15         & \multicolumn{1}{c|}{11.69}         & 13.52                    \\
Nesti-Net~\cite{ben2019nesti}                                       & 8.43          & 10.72         & 17.56          & \multicolumn{1}{c|}{22.63}          & 10.20         & \multicolumn{1}{c|}{10.66}         & 13.37                    & 8.43          & 10.57         & 15.00          & \multicolumn{1}{c|}{18.16}          & 10.20         & \multicolumn{1}{c|}{10.66}         & 12.17                    \\
DeepFit~\cite{ben2020deepfit}                                         & 6.51          & 9.21          & 16.73          & \multicolumn{1}{c|}{23.12}          & 7.93          & \multicolumn{1}{c|}{7.31}          & 11.80                    & 6.51          & 8.98          & 13.98          & \multicolumn{1}{c|}{19.00}          & 7.93          & \multicolumn{1}{c|}{7.31}          & 10.62                    \\
AdaFit~\cite{zhu2021adafit}                                        & 5.21          & 9.05          & 16.44          & \multicolumn{1}{c|}{21.94}          & 6.01          & \multicolumn{1}{c|}{5.90}          & 10.76                    & 5.21          & 8.79          & 13.55          & \multicolumn{1}{c|}{17.31}          & 6.01          & \multicolumn{1}{c|}{5.90}          & 9.46                    \\
GraphFit~\cite{li2022graphfit}                                        & 4.49          & 8.69          & 16.04 & \multicolumn{1}{c|}{21.64}          & 5.40          & \multicolumn{1}{c|}{5.20}          & 10.24                    & 4.49          & 8.43          & 13.00          & \multicolumn{1}{c|}{16.93}          & 5.40          & \multicolumn{1}{c|}{5.20}          & 8.91                     \\
HSurf-Net~\cite{li2022hsurf}                                       & 4.17          & 8.78          & 16.25          & \multicolumn{1}{c|}{21.61}          & 4.98          & \multicolumn{1}{c|}{4.86}          & 10.11                    & 4.17          & 8.52          & 13.23          & \multicolumn{1}{c|}{16.72}          & 4.98          & \multicolumn{1}{c|}{4.86}          & 8.75                     \\
Du~\etal~\cite{du2023rethinking}                                       & 4.11           & 8.66          & \textbf{16.02}          & \multicolumn{1}{c|}{21.57}          & 4.89          & \multicolumn{1}{c|}{4.83}          & 10.01                    &4.11           & 8.43          & 13.10          & \multicolumn{1}{c|}{17.08}          & 4.89          & \multicolumn{1}{c|}{4.83}          & 8.74                     \\
SHS-Net~\cite{li2023shs}                                         & 3.95          & 8.55          & 16.13          & \multicolumn{1}{c|}{\textbf{21.53}} & 4.91          & \multicolumn{1}{c|}{4.67}          & 9.96                     & 3.95          & 8.29          & 13.13          & \multicolumn{1}{c|}{16.60}          & 4.91          & \multicolumn{1}{c|}{4.67}          & 8.59                     \\
\cellcolor{boxbody}{Ours}                                            & \cellcolor{boxbody}{\textbf{3.86}} & \cellcolor{boxbody}{\textbf{8.45}} & \cellcolor{boxbody}{16.08}          & \cellcolor{boxbody}{21.89}          & \cellcolor{boxbody}{\textbf{4.85}} & \cellcolor{boxbody}{\textbf{4.45}} & \cellcolor{boxbody}{\textbf{9.93}}            & \cellcolor{boxbody}{\textbf{3.86}} & \cellcolor{boxbody}{\textbf{8.13}} & \cellcolor{boxbody}{\textbf{12.55}} & \cellcolor{boxbody}{\textbf{16.23}} & \cellcolor{boxbody}{\textbf{4.85}} & \cellcolor{boxbody}{{\textbf{4.45}}} & \cellcolor{boxbody}{\textbf{8.35}}\\
\hline
\end{tabular}}
\label{tab: pcpnet}
\end{table*}

\begin{figure*}[h]
	\centering
\includegraphics[scale=0.13]{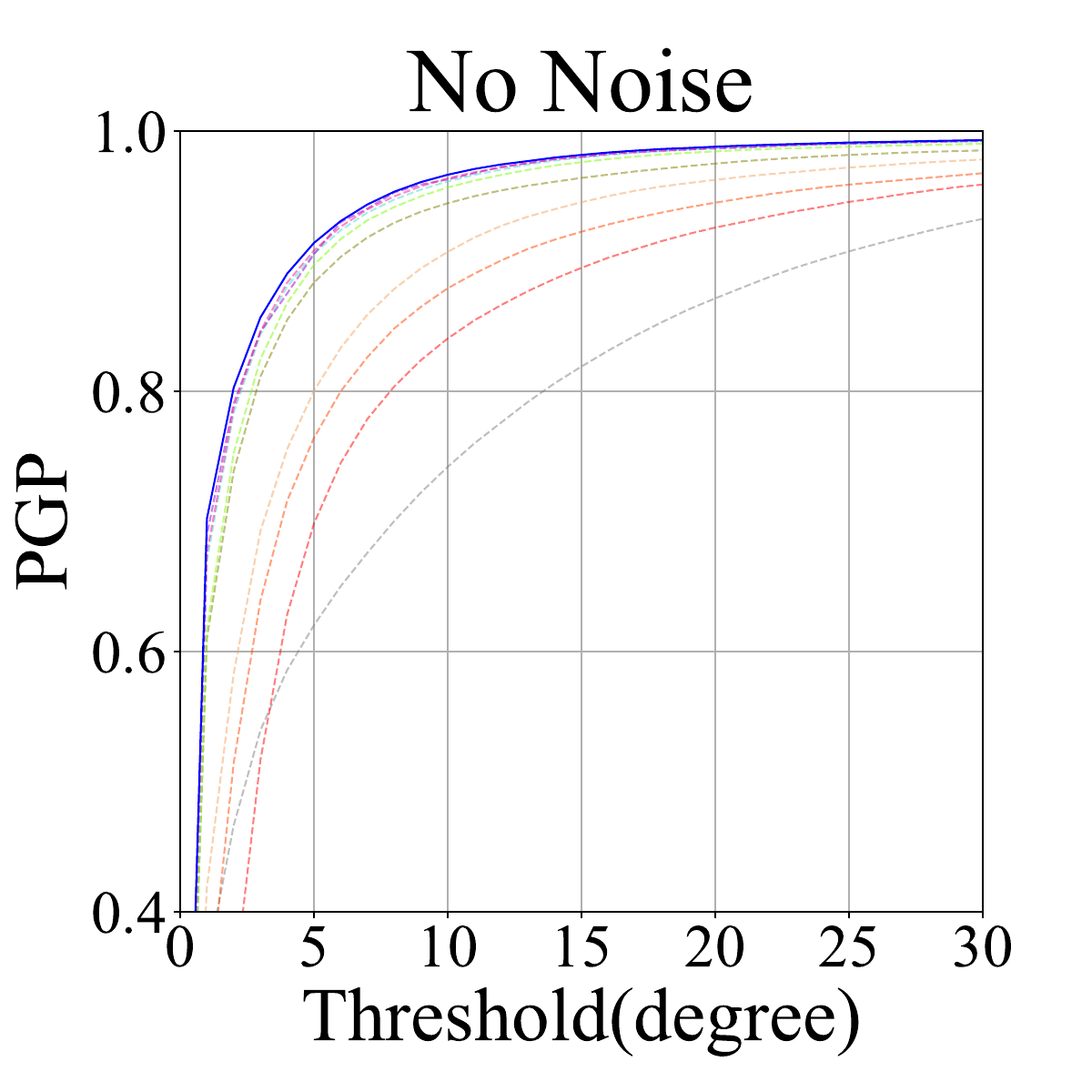}
\includegraphics[scale=0.13]{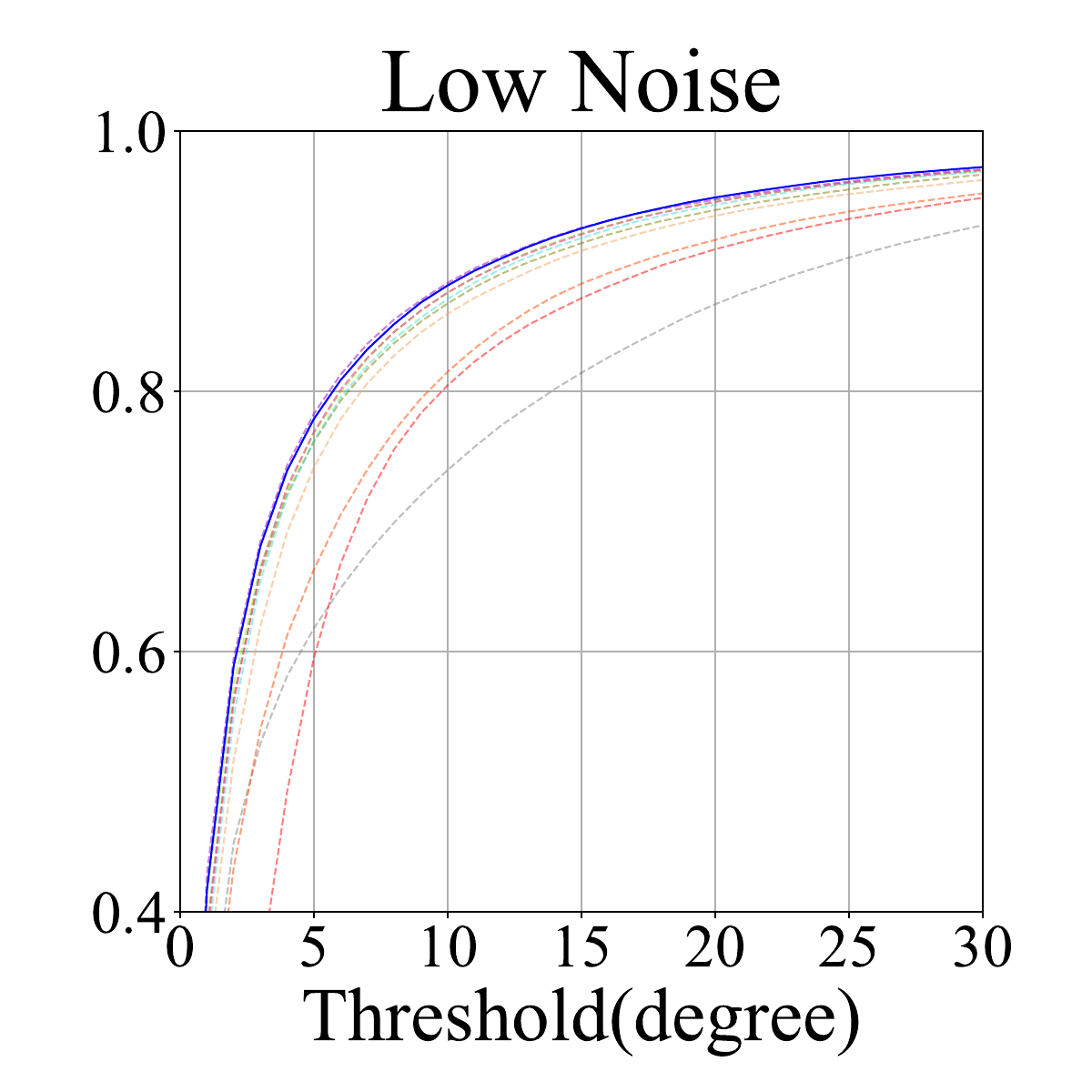}
\includegraphics[scale=0.13]{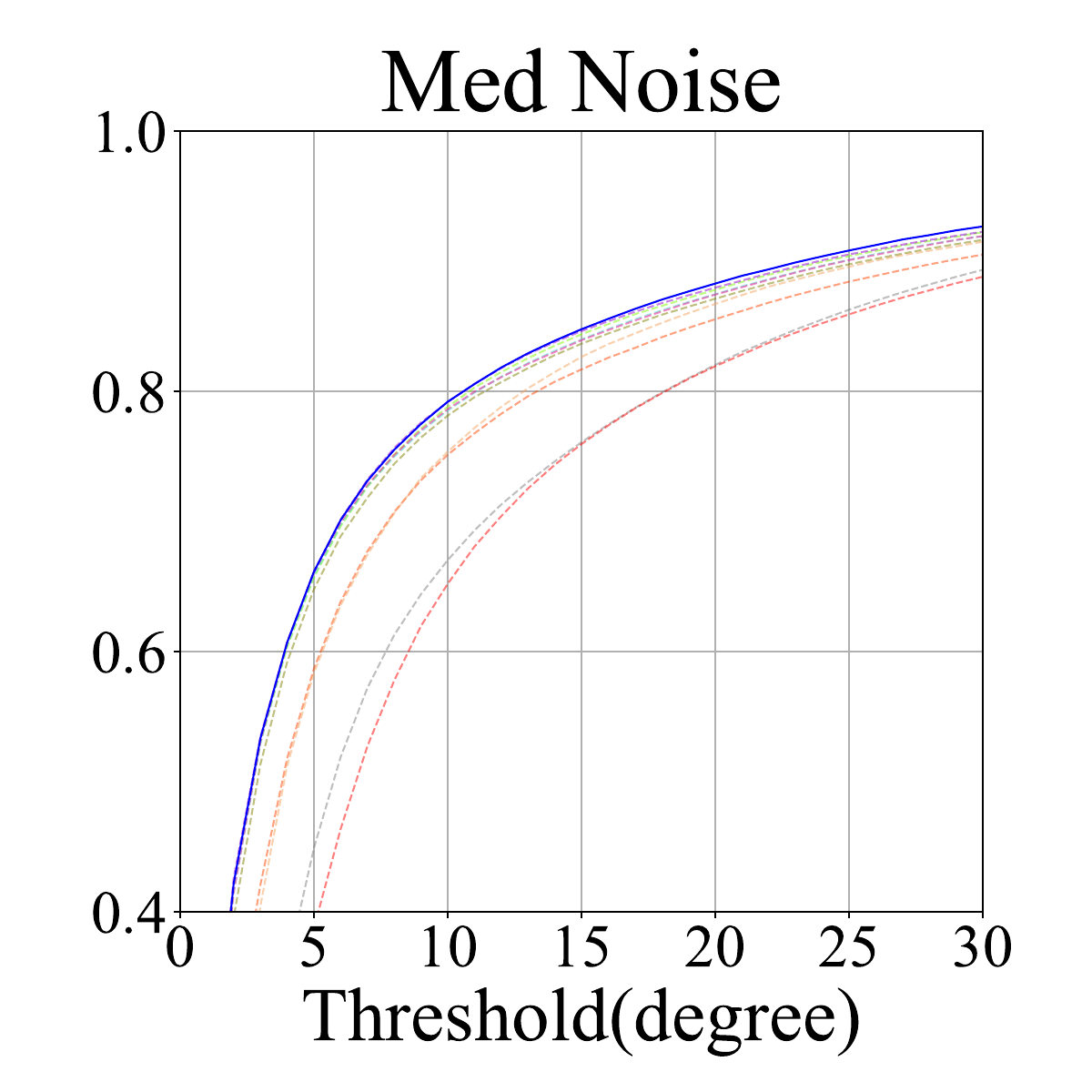}
\includegraphics[scale=0.13]{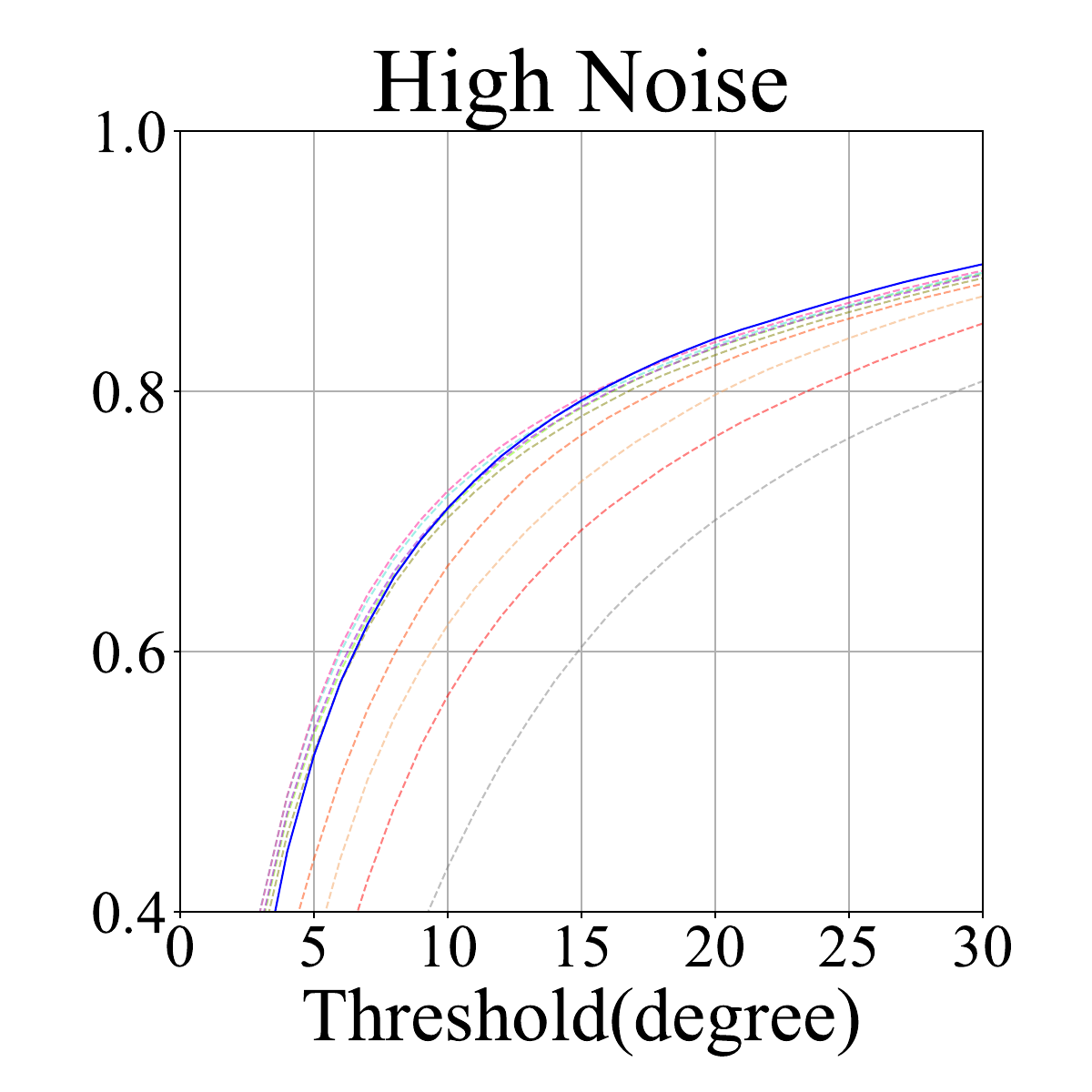}
\includegraphics[scale=0.13]{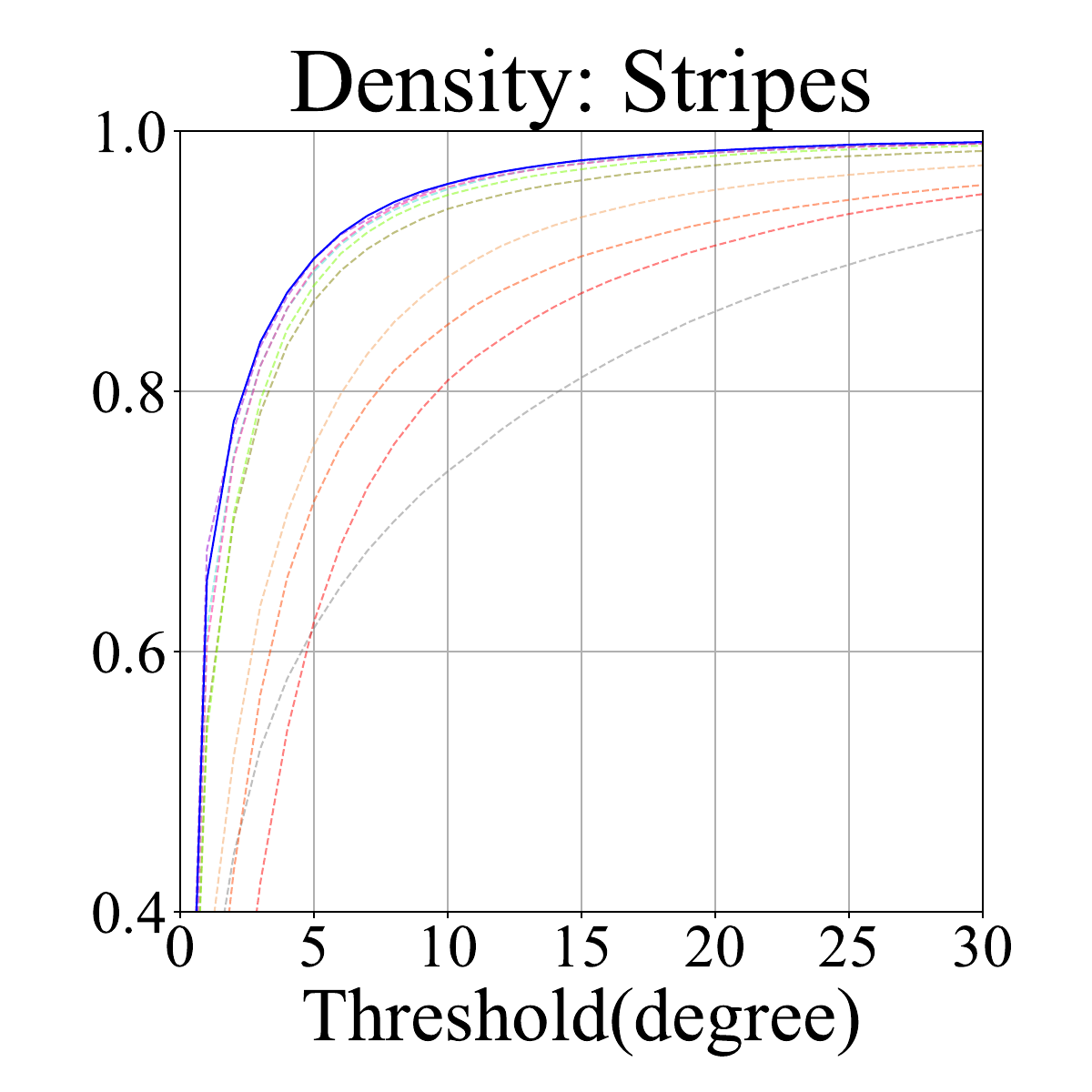}
\includegraphics[scale=0.13]{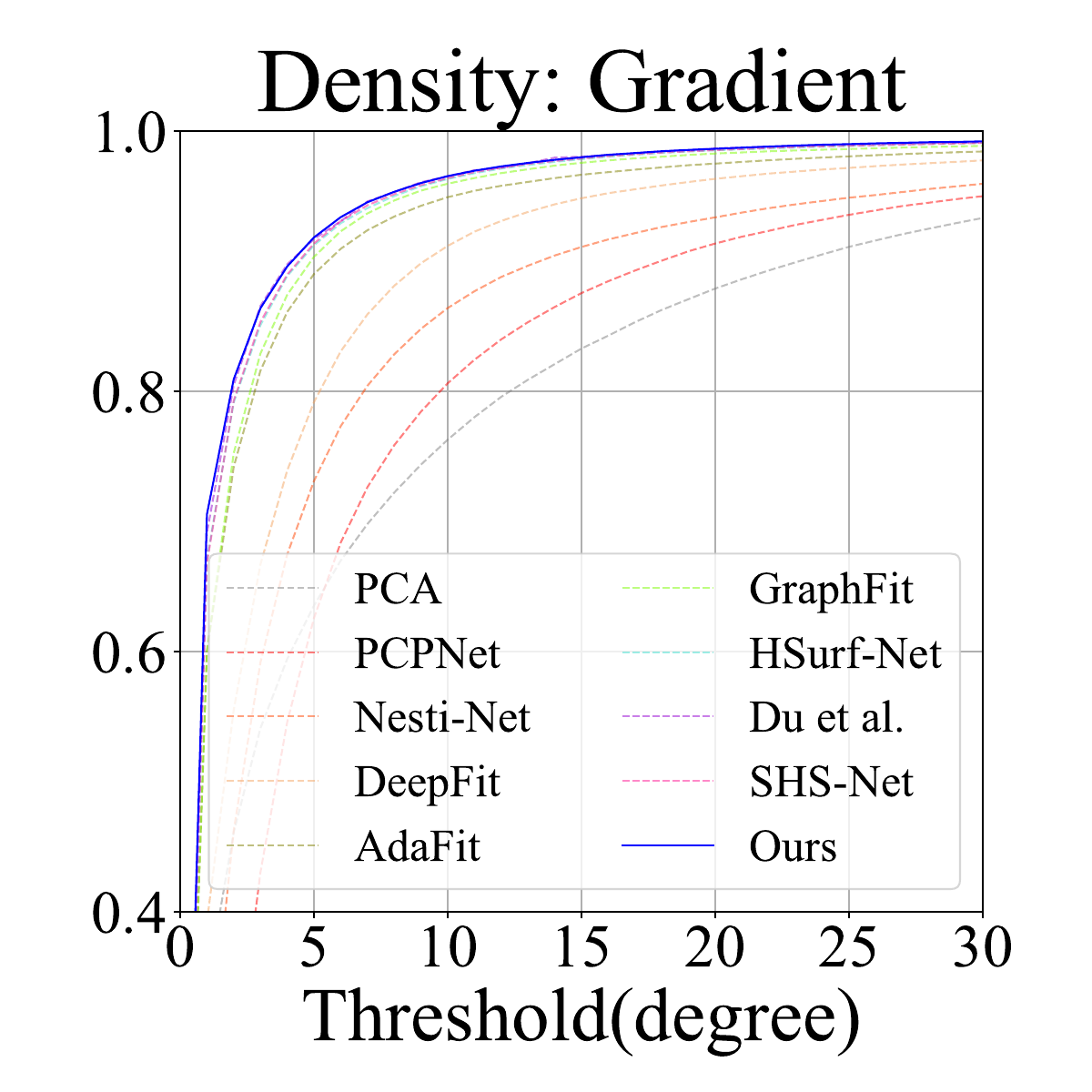}
\caption{AUC on the PCPNet dataset. $X$ and $Y$ axes are the angle threshold and the percentage of good point (PGP) normals.}
\label{fig: auc}
\end{figure*}

Previous methods group the local features by $k$-NN and capture the geometric information by MLP and maxpooling~\cite{li2022hsurf}. However, this manner often suffers from scale ambiguity and results in unsatisfactory robustness against noise. To solve this issue,  as presented in Fig.~\ref{fig: architecture_of_CMG-Net}(b), we construct graphs by $k$-NN with small and large scales and employ the skip-connection and maxpooling to capture the local structures. The \emph{Local Feature Extraction} (LFE) can be formulated as
\begin{equation}
\fontsize{6.5pt}{\baselineskip}\selectfont
\boldsymbol{f}_{i}^{n+1}\!=\!\mathrm{MaxPool}\left\{\phi_1\left(\varphi_1\left(\boldsymbol{f}_{i}^{n}\right), \varphi_1\left(\boldsymbol{f}_{i, j}^{n}\right), \varphi_1\left(\boldsymbol{f}_{i}^{n}-\boldsymbol{f}_{i, j}^{n}\right)\right)\right\}_{j=1}^{s_l},
\end{equation}
where $\boldsymbol{f}_{i, j}^{n}$ is the neighbor feature of the feature $\boldsymbol{f}_{i}^{n}$, $\varphi_1$ is the MLP layer, $\phi_1$ is the skip-connection layer, and $s_l$ represents the scale of $k$-NN with $l=1, 2$ in default. Moreover, we use an \emph{Attentional Feature Fusion} (AFF) architecture to aggregate the features which can benefit both the small and large scales. The AFF can be formulated as
\begin{equation}
\fontsize{6.5pt}{\baselineskip}\selectfont
M\left(\boldsymbol{f}_{i}^{s_1}, \boldsymbol{f}_{i}^{s_2}\right)=\mathrm{sigmoid}\left(\varphi_2\left(\mathrm{AvgPool}\left\{\boldsymbol{f}_{i}^{s_1}+\boldsymbol{f}_{i}^{s_2}\right\}_{i=1}^{N}\right)\right),
\end{equation}
\begin{equation}
\fontsize{6.5pt}{\baselineskip}\selectfont
\boldsymbol{f}_{i}=\varphi_3\left(\boldsymbol{f}_{i}^{s_1}\cdot M\left(\boldsymbol{f}_{i}^{s_1}, \boldsymbol{f}_{i}^{s_2}\right)+\boldsymbol{f}_{i}^{s_2}\cdot \left(1-M\left(\boldsymbol{f}_{i}^{s_1}, \boldsymbol{f}_{i}^{s_2}\right)\right)\right),
\end{equation}
where $\boldsymbol{f}_{i}^{s_1}$ abd $\boldsymbol{f}_{i}^{s_2}$ are the local structures with different scales of feature $\boldsymbol{f}_{i}$, $\varphi_2$ and $\varphi_3$ are the MLP layers, $N$ represents the cardinality of the input point cloud patch.

\subsubsection{Hierarchical Geometric Information Fusion.}
Recent approaches have proven the effectiveness of multi-scale global feature extraction~\cite{qi2017pointnet++, li2022hsurf, qin2022geometric}, however, large scale global information and local structures may be lost after point cloud downsampling. To alleviate this problem, as shown in Fig.~\ref{fig: architecture_of_CMG-Net}(c), we propose a \emph{hierarchical architecture} that combines the multi-scale global features with the local structures. During the Hierarchical Geometric Information Fusion, the global feature $\boldsymbol{G}_{N_h}$ of current scale $N_h$ can be formulated as
\begin{equation}
\boldsymbol{G}_{N_h}=\varphi_5\left(\mathrm{MaxPool}\left\{\varphi_4\left(\boldsymbol{f}_{i}^{N_h}\right)\right\}_{i=1}^{N_h}\right),
\end{equation}
where $\varphi_4$ and $\varphi_5$ are MLP layers. Meanwhile, the local structures $\boldsymbol{g}_{i}^{N_{h+1}}$ are captured by
\begin{scriptsize}
\begin{equation}
\boldsymbol{g}_{i}^{N_{h+1}}=\mathrm{MaxPool}\left\{\varphi_6\left(\boldsymbol{g}_{i, j}^{N_{h}}\right)\right\}_{j=1}^{s}+\boldsymbol{g}_{i}^{N_{h}}, i=1, ..., N_{h+1},
\end{equation}
\end{scriptsize}where $\boldsymbol{g}_{i, j}^{N_{h}}$ is the neighborhood feature of point $\boldsymbol{p}_i$ in the scope of the scale $N_{h+1}$, $s$ is the scale of the neighborhood features, and $\varphi_6$ represents the MLP layer. Then, we downsample the patch by decreasing the patch size. Moreover, we integrate the global features of the  current scale and the last scale with the local structures by
\begin{scriptsize}
\begin{equation}
\boldsymbol{f}_{i}^{N_{h+1}}=\varphi_7\left(\boldsymbol{G}_{N_h}, \boldsymbol{G}_{N_{h-1}}, \boldsymbol{g}_{i}^{N_{h+1}}\right) + \boldsymbol{f}_{i}^{N_h}, i=1, ..., N_{h+1},
\end{equation}
\end{scriptsize}where $\varphi_7$ is the MLP layer, and $N_{h+1}\leq N_{h}\leq N_{h-1}$.

\subsubsection{Decoder.}
Note that the point coordinates are important basic attributes for point cloud processing and the spatial relationship between them such as distance can guide the inference process of the network~\cite{zhao2021point,zhang2022geometry}. To explore this idea, we introduce two modules including \emph{Position Feature Fusion} (PFF) and \emph{Weighted Normal Prediction} (WNP) into the decoder part. As shown in Fig.~\ref{fig: architecture_of_CMG-Net}(d), during the PFF, we embed the neighborhood coordinates of each point and fuse them with the extracted feature by skip-connections, which can be formulated as
\begin{scriptsize}
\begin{equation}
\boldsymbol{F}_{i}=\mathrm{MaxPool}\left\{\phi_2\left(\boldsymbol{f}_{i}, \boldsymbol{p}_{i, j}-\boldsymbol{p}_{i}, \varphi_8\left(\boldsymbol{p}_{i, j}-\boldsymbol{p}_{i}\right)\right)\right\}_{j=1}^{s},
\end{equation}
\end{scriptsize}where $\boldsymbol{p}_{i, j}$ is the neighbor coordinate of the point $\boldsymbol{p}_{i}$, $\boldsymbol{f}_{i}$ is the extracted feature of $\boldsymbol{p}_{i}$, $s$ represents the neighborhood scale, $\varphi_8$ is the MLP layer and $\phi_2$ is the skip-connection. As shown in Fig.~\ref{fig: architecture_of_CMG-Net}(e), we predict weights based on the geometry information of each point and use the weighted features to predict the normal vector of the query point:
\begin{scriptsize}
\begin{equation}
\boldsymbol{n}=\varphi_{11}\left(\mathrm{MaxPool}\left\{\varphi_{10}\left(\boldsymbol{F}_{i}\cdot \mathrm{softmax}_{M}\left(\varphi_{9}\left(\boldsymbol{F}_{i}\right)\right)\right)\right\}_{i=1}^{M}\right),
\end{equation}
\end{scriptsize}where $\varphi_9$, $\varphi_{10}$ and $\varphi_{11}$ are the MLP layers, and the normalized $\boldsymbol{n}$ is the finally predicted unit normal vector.

\subsubsection{Loss function.}
To bridge the gap between the annotated normal and the noise-caused neighborhood geometry variation of the query point, we reformulate the sine loss by CND, namely, taking the normal $\boldsymbol{n}_{\boldsymbol{\tilde{p}}}$ of the nearest neighbor point $\boldsymbol{\tilde{p}}$ in the corresponding noise-free point cloud $\tilde{\mathcal{P}}$ as the ground truth
\begin{equation}
\mathcal{L}_1=\left\Vert\boldsymbol{n}_{\boldsymbol{\tilde{p}}}\times\boldsymbol{\hat{n}}_{\boldsymbol{p}}\right\Vert.
\end{equation}
Meanwhile, we use the transformation regularization loss and the z-direction transformation loss to constrain the output rotation matrix $\boldsymbol{R}\in\mathbb{R}^{3\times 3}$ of the QSTN~\cite{du2023rethinking}
\begin{equation}
\mathcal{L}_2=\left\Vert\boldsymbol{I}-\boldsymbol{R}\boldsymbol{R}^\mathrm{T}\right\Vert^2,
\end{equation}
\begin{equation}
\mathcal{L}_3=\left\Vert\boldsymbol{n}_{\boldsymbol{\tilde{p}}}\boldsymbol{R}\times{\boldsymbol{z}}\right\Vert,
\end{equation}
where $\boldsymbol{I}\in\mathbb{R}^{3\times 3}$ represents the identity matrix, $\boldsymbol{z}=\left(0, 0, 1\right)$. Additionally, to make full use of the spatial relationships between data points, we adopt the weight loss similar to Zhang~\etal~\shortcite{zhang2022geometry}
\begin{equation}
\mathcal{L}_4=\frac{1}{M}\sum_{i=1}^{M}(w_i-\hat{w}_i)^2,
\end{equation}
where $\hat{w}$ are the predicted weights for each data point, $M$ represents the cardinality of the downsampled patch, $w_i=\exp(-\left(\boldsymbol{p}_i\cdot\boldsymbol{n}_{\boldsymbol{\tilde{p}}}\right)^2/\delta^2)$ and $\delta=\max\left(0.05^2,0.3\sum_{i=1}^M\left(\boldsymbol{p}_i\cdot\boldsymbol{n}_{\boldsymbol{\tilde{p}}}\right)^2/M\right)$, where $\boldsymbol{p}_i$ is the point in the downsampled patch. Therefore, our final loss function is defined as
\begin{equation}
\mathcal{L}=\lambda_1\mathcal{L}_1+\lambda_2\mathcal{L}_2+\lambda_3\mathcal{L}_3+\lambda_4\mathcal{L}_4,
\end{equation}
where $\lambda_1=0.1$, $\lambda_2=0.1$, $\lambda_3=0.5$, and $\lambda_4=1$ are weighting factors.

\begin{table}[t]
\centering
\caption{Quantitative comparisons of CND on the PCPNet dataset with gradually increased noise.}
\resizebox{\linewidth}{!}{
\begin{tabular}{l|ccccc|c}
\hline
\multicolumn{1}{l|}{\multirow{2}{*}{{Method}}} & \multicolumn{5}{c|}{{Noise ($\sigma$)}}                                                       & \multirow{2}{*}{{Ave.}} \\
\multicolumn{1}{c|}{}                          & 0.125\%        & 0.25\%         & 0.5\%          & 0.75\%         & 1.25\%         &                         \\ \hline
PCA                                            & 14.46          & 15.09          & 17.75          & 21.40          & 31.81          & 20.10                   \\
n-jet                                          & 14.40          & 14.98          & 17.67          & 21.50          & 32.08          & 20.12                   \\
PCPNet                                         & 13.42          & 15.52          & 18.12          & 20.02          & 23.92          & 18.20                   \\
Nesti-Net                                      & 13.34          & 14.33          & 16.63          & 18.34          & 22.31          & 16.99                   \\
DeepFit                                        & 11.70          & 12.83          & 15.62          & 17.64          & 24.01          & 16.36                   \\
AdaFit                                         & 11.42          & 12.95          & 15.52          & 17.02          & 21.74          & 15.73                   \\
GraphFit                                       & 11.01          & \textbf{12.40} & 15.05          & 16.66          & 20.56          & 15.14                   \\
HSurf-Net                                      & 11.04          & 12.67          & 15.33          & 16.79          & 20.67          & 15.30                   \\
Du~\etal~                                      & 10.97          & 12.48          & 15.17          & 16.77          & 21.04          & 15.29                   \\
SHS-Net                                        & 10.90          & 12.66          & 15.18          & 16.59          & 20.89          & 15.24                   \\
\cellcolor{boxbody}{Ours}                                           & \cellcolor{boxbody}{\textbf{10.60}} & \cellcolor{boxbody}{12.56}          & \cellcolor{boxbody}{\textbf{14.89}} & \cellcolor{boxbody}{\textbf{16.31}} & \cellcolor{boxbody}{\textbf{19.62}} & \cellcolor{boxbody}{\textbf{14.79}}          \\ \hline
\end{tabular}}
\label{tab: pcpnet_nosie}
\end{table}

\begin{figure}[t]
\centering
\includegraphics[scale=0.35]{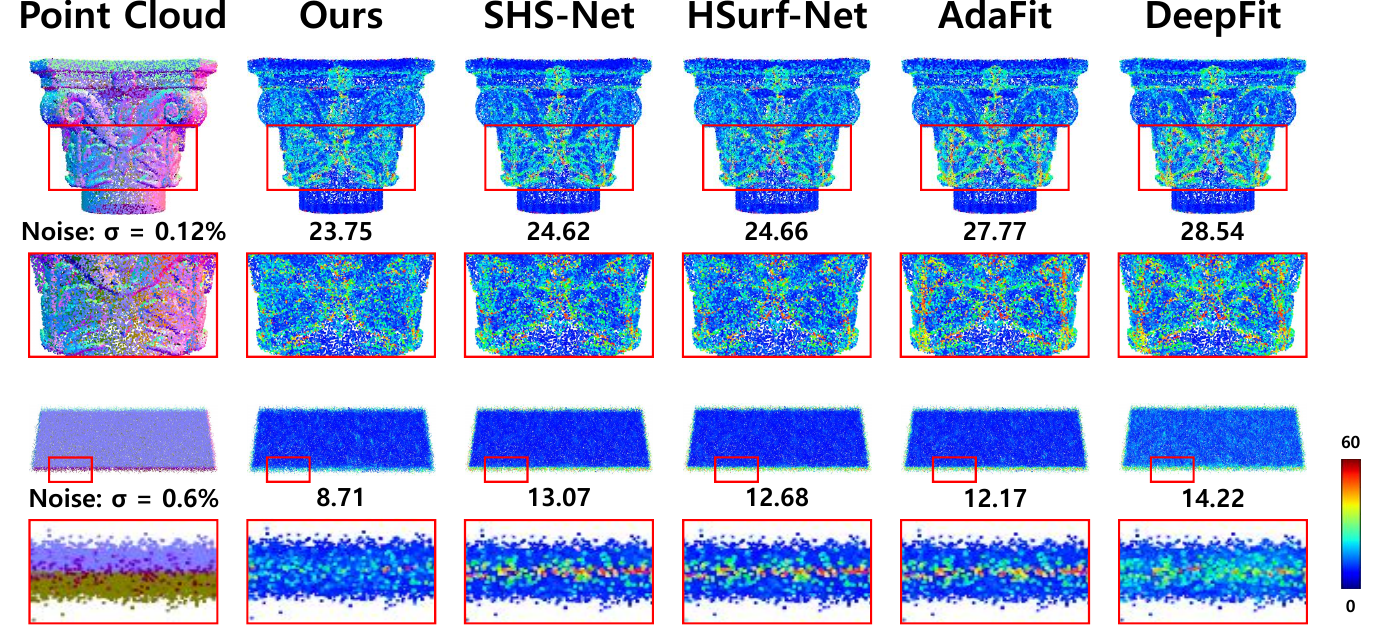}
\caption{Comparisons on the PCPNet datasts (Noise: $\sigma=0.12\%, 0.6\%$). We use the heat map to visualize the CND  error. }
\label{fig: pcpnet_examples}
\end{figure}

\section{Experimental Results}

\subsubsection{Datasets.}
As predecessor approaches, we first adopt the synthetic dataset PCPNet~\cite{guerrero2018pcpnet} for comparison, in which we follow the same experimental setups including train-test split, adding noise, and changing distribution density on test data. To test the generalization capability of our method, we then evaluate and compare the models trained on the PCPNet on the real-world indoor SceneNN dataset~\cite{hua-pointwise-cvpr18}.

\subsubsection{Implementation details.}
We set the input patch size $N=700$ and the downsampling factors $\rho=\{2/3, 2/3, 2/3, 1\}$. The scales of $k$-NN in the LFE are equivalent to $16$ and $32$, and $s=\{32, 32, 16, 16\}$ in the Hierarchical Geometric Information Fusion. The number of the neighbor points during the PPF is $16$. We adopt the AdamW~\cite{loshchilov2017decoupled} optimizer with initial learning rate $5 \times 10^{-4}$ for training. The learning rate is decayed by a cosine function. Our model is trained with a 64 batch size on an NVIDIA A100 GPU in 900 epochs. More implementation details are reported in \emph{Supplementary Materials (SM)}.

\subsubsection{Evaluation.}
We adopt the proposed CND metric to assess the normal estimation results and compare it with the RMSE. Moreover, we use the \emph{Area Under the Curve} (AUC) metric to analyze the error distribution of the predicted normals. AUC is attained by the \emph{Percentage of Good Points} (PGP) metric, which measures the percentage of normal vectors with errors below different angle thresholds.

\subsection{Results on Synthetic Data}

\subsubsection{PCPNet.}
Table~\ref{tab: pcpnet} reports the statistical results of all compared approaches on the PCPNet dataset, measured in terms of both RMSE and CND metrics. As observed, our method achieves the overall highest normal estimation accuracy across different scenarios, particularly in scenarios with noise. In comparison to RMSE, the CND metric allows for more accurate and faithful prediction evaluations while mitigating the annotation inconsistency. Additionally, the AUC results of the CND metric are illustrated in Fig.~\ref{fig: auc}, where our method still showcases the superior performance, suggesting its remarkable stability across different angular thresholds. Qualitative comparison results are presented in Fig.~\ref{fig: pcpnet_examples}. Notably, our method exhibits the smallest errors in regions characterized by noise and intricate geometry.

\begin{table}[t]
\centering
\caption{Statistical CND results on the SceneNN dataset.}
\resizebox{\linewidth}{!}{
\begin{tabular}{l|lllllll}
\hline
{Method}  & Ours           & SHS-Net & Du~\etal~& HSurf-Net     & GrapFit & AdaFit & DeepFit \\ \hline
Clean   & 6.92           & 7.20    & 6.97  & \textbf{6.73} & 7.38    & 7.55   & 9.46    \\
Noise   & \textbf{10.82} & 11.30   & 10.94 & 11.30         & 11.38   & 11.82  & 12.27   \\
Ave. & \textbf{8.87}  & 9.25    & 8.96  & 9.02          & 9.38    & 9.97   & 10.86   \\ \hline
\end{tabular}}
\label{tab: scenennn}
\end{table}

\begin{figure}[t]
\centering
\includegraphics[scale=0.30]{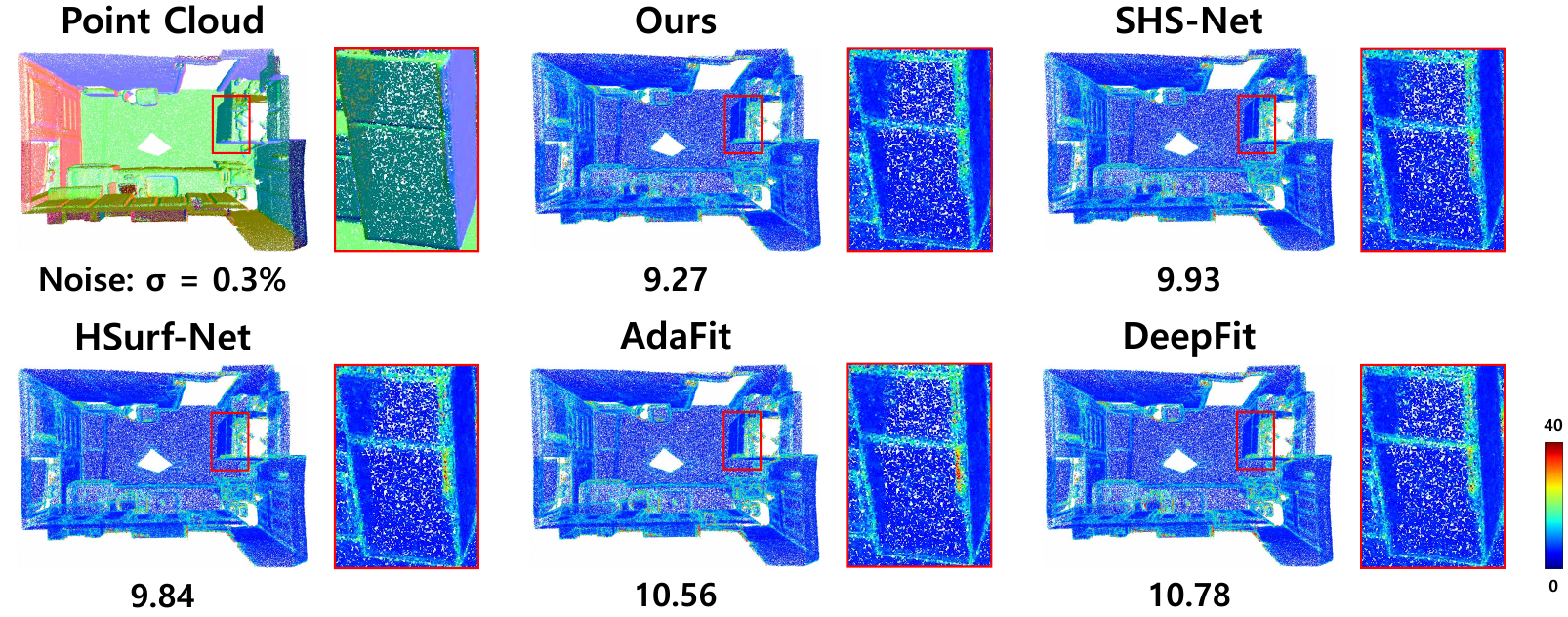}
\caption{Qualitative comparisons on the SceneNN datasets (Noise: $\sigma=0.3\%$). }
\label{fig: scenenn_examples}
\end{figure}

\subsubsection{Robustness to noise.}
Subsequently, we specifically employ five representative models from the PCPNet dataset to assess the robustness against noise. We introduce varying levels of noise to these data which encompass one CAD model and four scanned point clouds. The quantitative outcomes displayed in Table~\ref{tab: pcpnet_nosie} indicate that our method exhibits superior performance compared to competitors, particularly in scenarios contaminated by high levels of noise.

\subsection{Generalization to Real-world Data}
Next, we investigate the generalization capability using the real-world indoor SceneNN dataset. Results in Table~\ref{tab: scenennn} suggest that our method has the highest normal estimation accuracy in an average sense. The qualitative results presented in Fig.~\ref{fig: scenenn_examples} exhibit our superiority. It is noticeable that our method successfully preserves more geometric details, such as the handle of the refrigerators. Additionally, more results on different real-word datasets can be found in \emph{SM}.

\subsection{Ablation Study}

\subsubsection{CND-modified loss function.}To demonstrate the effectiveness and generalization of the newly introduced  CND-modified loss function, we conduct experiments on the PCPNet dataset, comparing the results with and without its incorporation. We employ representative methods, including the deep surface fitting method DeepFit~\cite{ben2020deepfit}, as well as the regression methods PCPNet~\cite{guerrero2018pcpnet}, Hsurf-Net~\cite{li2022hsurf}, and Ours. Table~\ref{tab: cnd} highlights the impact of the CND component, demonstrating its significant enhancement in normal estimation accuracy for both deep surface fitting and regression methods.

\begin{table}[t]
\centering
\caption{Network training with or without the CND-modified loss function on the PCPNet dataset.}
\resizebox{\linewidth}{!}{
\begin{tabular}{l|cc|cc|cc|cc}
\hline
Method     & \multicolumn{2}{c|}{Ours} & \multicolumn{2}{c|}{HSurf-Net} & \multicolumn{2}{c|}{DeepFit} & \multicolumn{2}{c}{PCPNet} \\ \hline
$\mathcal{L}_{\mathrm{CND}}$& \checkmark                &        & \checkmark                  &           & \checkmark                 &          & \checkmark                &         \\ \hline
No Noise   & 3.86             & 3.85   & 4.24               & 4.17      & 6.53              & 6.51     & \textbf{8.52}    & 9.62    \\
Noise: ($\sigma=0.12\%$)  & \textbf{8.13}    & 8.23   & \textbf{8.50}      & 8.52      & \textbf{8.77}     & 8.98     & \textbf{10.40}   & 11.23   \\
Noise: ($\sigma=0.6\%$)  & \textbf{12.55}   & 12.76  & \textbf{12.83}     & 13.22     & \textbf{13.66}    & 13.98    & \textbf{15.71}   & 17.28   \\
Noise: ($\sigma=1.2\%$) & \textbf{16.23}   & 16.46  & \textbf{16.47}     & 16.71     & \textbf{18.69}    & 19.00    & \textbf{18.31}   & 20.16   \\
Density: Stripes    & 4.85             & 4.65   & 5.18               & 4.98      & 7.95              & 7.93     & \textbf{9.96}    & 11.15   \\
Density: Gradients  & \textbf{4.45}    & 4.51   & 4.94               & 4.86      & \textbf{7.31}     & 7.31     & \textbf{10.25}   & 11.69   \\ \hline
Ave.    & \textbf{8.35}    & 8.41   & \textbf{8.69}      & 8.75      & \textbf{10.48}    & 10.62    & \textbf{12.19}   & 13.52   \\ \hline
\end{tabular}}
\label{tab: cnd}
\end{table}

\subsubsection{Network architecture.}

\begin{table}[t]
\centering
\caption{Ablation studies with the (a) multi-scale local feature aggregation; (b) hierarchical architecture; (c) decoder; (d) QSTN and (e) scale selection and downsampling factor.}
\resizebox{\linewidth}{!}{
\begin{tabular}{ll|cccc|cc|c}
\hline
\multicolumn{2}{c|}{\multirow{2}{*}{Category}}                 & \multicolumn{4}{c|}{Noise ($\sigma$)}                                    & \multicolumn{2}{c|}{Density}  & \multirow{2}{*}{Ave.} \\
\multicolumn{2}{c|}{}                                          & None          & 0.12\%       & 0.6\%          & 1.2\%          & Stripes       & Gradient      &                         \\ \hline
\multirow{3}{*}{(a)} & w/o Local Feature Extration (LFE)             & 4.05          & 8.23          & 12.76          & 16.45          & 4.93          & 4.68          & 8.52                    \\
                     & w/ Single-scale Local Feature Extration & 3.98          & 8.20          & 12.76          & 16.40          & 4.96          & 4.66          & 8.50                    \\
                     & w/o Attentional Feature Fusion (AFF)    & 3.96          & 8.19          & 12.68          & 16.39          & 4.78          & 4.62          & 8.44                    \\ \hline
\multirow{3}{*}{(b)} & w/o Hierarchical Architecture           & 3.88          & 8.45          & 13.80          & 18.93          & 4.87          & 4.50          & 9.07                    \\
                     & w/o Muli-scale Global Feature           & 3.87          & 8.27          & 12.56          & 16.24          & 4.94          & \textbf{4.45} & 8.39                    \\
                     & w/o Local Feature                       & 3.98          & 8.46          & 12.68          & 16.21          & 5.00          & 4.60          & 8.49                    \\ \hline
\multirow{2}{*}{(c)} & w/o Position Feature Fusion (PFF)       & 3.93          & 8.15          & 12.62          & 16.24          & 4.87          & 4.68          & 8.42                    \\
                     & w/o Weighted Normal Prediction (WNP)    & 4.32          & 8.23          & 12.56          & 16.22          & 5.01          & 4.89          & 8.54                    \\ \hline
\multirow{2}{*}{(d)} & w/o QSTN                                & 4.04          & 8.34          & 12.67          & 16.41          & 4.95          & 4.74          & 8.53                    \\
                     & w/o  Z-direction Transformation Loss    & 4.02          & 8.18          & 12.62          & 16.32          & 4.98          & 4.72          & 8.47                    \\ \hline
\multirow{3}{*}{(e)} & $N=600$                                 & 3.90          & 8.36          & 12.59          & 16.35          & 4.78          & 4.46          & 8.41                    \\
                     & $N=800$                                 & 4.05          & 8.24          & 12.52          & \textbf{16.17} & 4.99          & 4.65          & 8.44                    \\
                     & $\rho=\{1/3, 1/3, 1, 1\}$               & 3.95          & 8.25          & 12.56          & 16.39          & \textbf{4.76} & 4.61          & 8.42                    \\
                     & $\rho=\{1/2, 1/2, 1, 1\}$               & 3.93          & 8.22          & 12.72          & 16.29          & 4.80          & 4.50          & 8.41                    \\ \hline
\multicolumn{1}{l}{} & Ours                                    & \textbf{3.86} & \textbf{8.13} & \textbf{12.55} & 16.23          & 4.85          & \textbf{4.45} & \textbf{8.35}           \\ \hline
\end{tabular}
}
\label{tab: ablation}
\end{table}

\begin{figure}[t]
\centering
\includegraphics[scale=0.4]{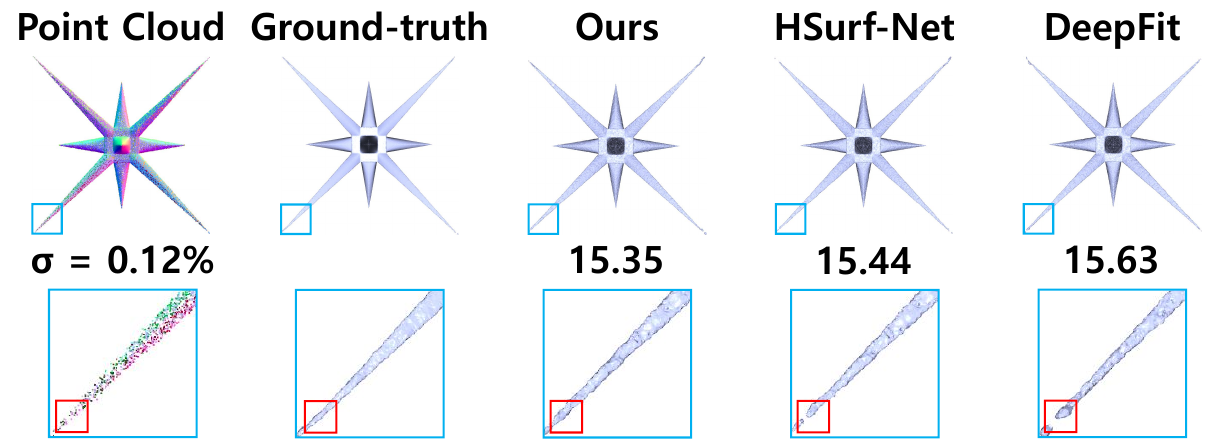}
\caption{Comparisons on Poisson surface reconstruction.}
\label{fig: reconstruction_examples}
\end{figure}

CMG-Net comprises three key components: Multi-scale Local Feature Aggregation, Hierarchical Geometric Information Fusion, and Decoder. We delve into the functions of them on the PCPNet dataset. \\
(1). In the Multi-scale Local Feature Aggregation, we capture the local structure using two scales and integrate them by AFF. Table~\ref{tab: ablation}(a) reports the results of 1) without LFE; 2) with single-scale LFE, and 3) integrating multi-scale local features directly by MLP instead of AFF. As observed, compared with Ours, the multi-scale local features with AFF can effectively improve the network performance.\\
(2). To validate the effectiveness of the Hierarchical Geometric Information Fusion, we carry out experiments using the model with a fixed global scale that is equivalent to the output scale of CMG-Net. Additionally, we compare the results of the models without the global feature of the last scale and the local feature in the hierarchical architecture. Results shown in Table~\ref{tab: ablation}(b) demonstrate that the Hierarchical Geometric Information Fusion operation can also boost the normal estimation performance.\\
(3). Table~\ref{tab: ablation}(c) shows the ablation studies of the Decoder part, suggesting the effectiveness of PFF and WNP.\\
(4). We also investigate the functionality of QSTN, the input patch sizes $N$, and the downsampling factors $\rho$ in Table~\ref{tab: ablation}(d) and Table~\ref{tab: ablation}(e), where quantitative results validate their usefulness in our method.

\subsection{Application of the Proposed Method}
We also demonstrate the application of our method on downstream tasks. Fig.~\ref{fig: reconstruction_examples} presents the \emph{Poisson surface reconstruction}~\cite{kazhdan2006poisson} results using the normal vectors predicted by competing approaches. Compared with ground-truth surfaces, our method achieves the best reconstruction quality (quantified by the \emph{Symmetric Mean Hausdorff Distance} (SMD)($\times10^{-4}$)), especially in shape details of noisy regions, underscoring the higher accuracy of our normal estimation. We provide more reconstruction instances and highlight the application of our newly developed method to point cloud denoising in the \emph{SM}.

\section{Limitations}
While our method has demonstrated remarkable normal estimation accuracy across diverse 3D models, especially in noisy scenarios, it is not yet real-time capable and still depends on annotated training data, as is the case with previous approaches. Therefore, it is highly desirable in the future to reduce the computation time and delve into semi-supervised or unsupervised normal estimation frameworks.

\section{Conclusions}
We propose a novel method for robust normal estimation in unorganized point clouds, which shows superiority across various datasets and scenarios. We identify the issue of direction inconsistency in predecessor approaches and introduce the CND metric to address this concern. This not only boosts the network training and evaluation, but also greatly enhances the network robustness against noisy disturbance. Additionally, we design an innovative architecture that combines multi-scale local and global feature extraction with hierarchical information fusion to deal with scale selection ambiguity. Extensive experiments validate that our method outperforms competitors in terms of both accuracy and robustness for normal estimation. Moreover, we demonstrate its ability to generalize in real-world settings and downstream application tasks.

\section*{Acknowledgements}
This work was partially funded by the National Natural Science Foundation of China (62172415, 12022117, 62102418),  the CAS Project for Young Scientists in Basic Research (YSBR-034), and the Beijing Science and Technology Plan Project (Z231100005923033).

\bibliography{aaai24}

\clearpage
\appendix

\begin{appendices}
\section*{Supplementary Materials}
In this exposition, we present more implementation details, experimental comparisons, and qualitative results to support our work. Concretely, the following content comprising of
\begin{enumerate}
\item implementation details with respect to the local features in Hierarchical Geometric Information Fusion and the QSTN structure,
\item explanation of the difference between CND and Mean Square Angular Error (MSAE)\cite{lu2020deep},
\item thorough comparisons with the results of normal estimation using denoising pre-processing,
\item more quantitative and qualitative results on {real-word datasets} and {point clouds contaminated by heavier noise},
\item employment of our method to real-time application and downstream tasks containing \emph{surface reconstruction} and \emph{denoising}
\end{enumerate} are reported.

\section{More Implementation Details}

In this section, we provide more implementation details of our method, especially the local features in  Hierarchical Geometric Information Fusion and the QSTN structure are reported.

\subsection{Local Features in Hierarchical Architecture}

To capture the local structures of both geometry and semantics in the Hierarchical Geometric Information Fusion module, we leverage various local features in the odd and even hierarchical layers. The local features of the odd hierarchical layers are defined by
\begin{equation}
\label{equ: local_feature1}
\boldsymbol{g}_{i, j}^{o}=\text{Concat}\left(\boldsymbol{p}_{i}, \boldsymbol{p}_{i}-\boldsymbol{p}_{i, j}, \varphi\left(\boldsymbol{p}_{i}-\boldsymbol{p}_{i, j}\right)\right),
\end{equation}
where $\boldsymbol{p}_{i, j}$ is the neighbor coordinate of the point $\boldsymbol{p}_{i}$ and $\varphi$ represents the MLP layer. Concurrently, the even ones put more focus on semantic feature defined as
\begin{equation}
\label{equ: local_feature2}
\boldsymbol{g}_{i, j}^{e}=\text{Concat}\left(\boldsymbol{p}_{i}, \boldsymbol{p}_{i}-\boldsymbol{p}_{i, j}, \boldsymbol{f}_{i}-\boldsymbol{f}_{i, j}\right),
\end{equation}
where $\boldsymbol{f}_{i}$ and $\boldsymbol{f}_{i, j}$ are the semantic features of $\boldsymbol{p}_{i}$ and its neighborhood.

In addition, to validate the effectiveness of our newly proposed feature extraction method, we conduct ablation studies on the models
\begin{enumerate}
    \item without local features;
    \item with local feature in Eq.~\ref{equ: local_feature1} only;
    \item with local feature in Eq.~\ref{equ: local_feature2} only.
\end{enumerate}
The results on the PCPNet dataset presented in Table~\ref{tab: ablation studies on hierarchical architecture} demonstrate that different local features in the odd and even hierarchical layers can significantly enhance the performance on normal estimation.

\begin{table}[t]
\caption{Ablation studies on the local features in the Hierarchical Geometric Information Fusion.}
\resizebox{\linewidth}{!}{\begin{tabular}{c|cccc}
\hline
Method     & Ours           & (3)                        & (2)                        & (1)               \\ \hline
No Noise   & \textbf{3.86}  & 3.91                       & 3.89                       & 3.98              \\
Noise: ($\sigma=0.12\%$) & \textbf{8.13}  & 8.22                       & 8.18                       & 8.46              \\
Noise: ($\sigma=0.6\%$)  & \textbf{12.55} & 12.64                      & 12.56                      & 12.68             \\
Noise: ($\sigma=1.2\%$) & \textbf{16.23} & 16.28                      & 16.25                      & 16.24             \\
Density: Stripes    & \textbf{4.85}  & 4.95                       & 4.93                       & 5.00              \\
Density: Gradients  & \textbf{4.45}  & 4.62                       & 4.71                       & 4.60              \\
Average    & \textbf{8.35}  & 8.44                       & 8.42                       & 8.49              \\ \hline
\end{tabular}}
\label{tab: ablation studies on hierarchical architecture}
\end{table}

\subsection{QSTN Structure}

\begin{figure}[t]
\centering
\includegraphics[scale=0.35]{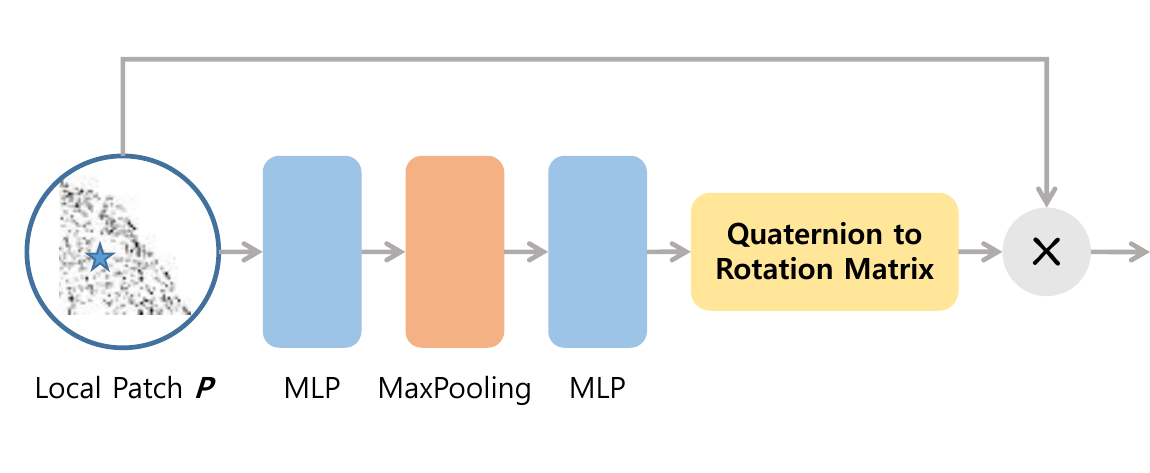}
\caption{Architecture of QSTN.}
\label{fig: qstn}
\end{figure}

Given a point cloud patch $\boldsymbol{P}=\{\boldsymbol{p}_i\in\mathbb{R}^3\}_{i=1}^N$, the QSTN part~\cite{qi2017pointnet,du2023rethinking} first computes the quaternion by MLP layers and then translates this quaternion into a rotation matrix $\boldsymbol{R}\in\mathbb{R}^{3\times 3}$, as shown in Fig.~\ref{fig: qstn}.

\section{More Normal Estimation Results}

In this section, we further explain the difference between CND and MSAE, a metric used to evaluate the denoising results of CAD models and compare our method with the normal estimator with denoising pre-processing~\cite{zhang2020pointfilter}. Additionally, more qualitative results on the point clouds contaminated by heavier noise and LiDAR datasets are provided.

\subsection{Differences between CND and MSAE Metrics} 
\begin{figure}[b]
\centering
\includegraphics[scale=0.5]{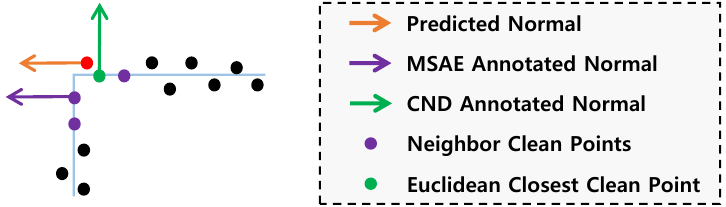}
\caption{Differences between the CND and MSAE metrics when wrong normal predictions on noisy edge points occur.}
\label{fig: comparison_between_the_MSAE_and _CND}
\end{figure}

As a metric for denoising, MSAE searches nearby ground-truth points’ normals and picks the minimal normal error. As shown in Fig.~\ref{fig: comparison_between_the_MSAE_and _CND}, our closest distance-induced CND metric effectively addresses direction inconsistency caused by noise, distinguishing it from MSAE, which primarily focuses on normal similarity and selects the neighboring point with the minimal normal error instead of relative coordinates. As shown in Table~\ref{tab: cnd vs msae}, we set the neighbor points size in MSAE to 4 and train the same network by the CND-loss and MSAE-loss respectively, while our CND-loss consistently attains superior results on both metrics in the test phase.

\begin{table}[t]
\centering
\caption{Test results of CND and MSAE on the PCPNet dataset.}
\renewcommand{\arraystretch}{1.3}
\resizebox{\linewidth}{!}{
\begin{tabular}{cl|ccccccc}
\hline
\multicolumn{2}{c|}{\multirow{3}{*}{Category}} & \multicolumn{7}{c}{MSAE}                                                                                                                                            \\ \cline{3-9}
\multicolumn{2}{c|}{}                          & \multicolumn{4}{c|}{Noise $\sigma$}                                                         & \multicolumn{2}{c|}{Density}                       & \multirow{2}{*}{Averge} \\
\multicolumn{2}{c|}{}                          & None          & 0.125\%       & 0.6\%          & \multicolumn{1}{c|}{1.2\%}          & Stripes       & \multicolumn{1}{c|}{Gradient}      &                         \\ \hline
\multicolumn{2}{c|}{L MSAE}                    & 4.12          & 6.69          & 9.41           & \multicolumn{1}{c|}{12.82}          & 5.45          & \multicolumn{1}{c|}{5.22}          & 7.29                    \\
\multicolumn{2}{c|}{L CND (ours)}              & \textbf{2.65} & \textbf{5.34} & \textbf{9.23}  & \multicolumn{1}{c|}{\textbf{12.74}} & \textbf{3.29} & \multicolumn{1}{c|}{\textbf{2.99}} & \textbf{6.04}           \\ \hline\hline
\multicolumn{2}{c|}{\multirow{3}{*}{Category}} & \multicolumn{7}{c}{CND}                                                                                                                                             \\ \cline{3-9}
\multicolumn{2}{c|}{}                          & \multicolumn{4}{c|}{Noise $\sigma$}                                                         & \multicolumn{2}{c|}{Density}                       & \multirow{2}{*}{Averge} \\
\multicolumn{2}{c|}{}                          & None          & 0.125\%       & 0.6\%          & \multicolumn{1}{c|}{1.2\%}          & Stripes       & \multicolumn{1}{c|}{Gradient}      &                         \\ \hline
\multicolumn{2}{c|}{L MSAE}                    & 4.66          & 10.62         & 13.37          & \multicolumn{1}{c|}{16.31}          & 5.63          & \multicolumn{1}{c|}{5.48}          & 9.35                    \\
\multicolumn{2}{c|}{L CND (ours)}              & \textbf{3.86} & \textbf{8.13} & \textbf{12.55} & \multicolumn{1}{c|}{\textbf{16.23}} & \textbf{4.85} & \multicolumn{1}{c|}{\textbf{4.45}} & \textbf{8.35}           \\ \hline
\end{tabular}}
\label{tab: cnd vs msae}
\end{table}

\begin{table}[t]
\caption{Results of the comparisons with the denosing pre-processing on the noisy part of the PCPNet Dataset.}
\resizebox{\linewidth}{!}{\begin{tabular}{c|ccc}
\hline
Method     & ours           & w/o $\mathcal{L}_{\mathrm{CND}}$ & w/ denoising \\ \hline
Noise: ($\sigma=0.12\%$)  & \textbf{8.13}  & 8.23     & 13.79        \\
Noise: ($\sigma=0.6\%$)  & \textbf{12.55} & 12.76    & 16.28        \\
Noise: ($\sigma=1.2\%$) & \textbf{16.23} & 16.46    & 17.91        \\
Average    & \textbf{12.30} & 12.48    & 15.99        \\ \hline
\end{tabular}}
\label{tab: comparison denoising}
\end{table}

\subsection{Comparisons with Denoising Pre-Processing}

To further prove the effectiveness of our proposed CND-Modified loss function, we conduct experiments on the models trained with or without CND-Modification as well as the model with denoising pre-processing. All of the compared models have the same architecture. The results in Table~\ref{tab: comparison denoising} indicate that due to the destruction of geometrical information and the oversmoothing of shape details, the denoising pre-processing decreases the accuracy of normal estimation instead. In contrast, the CND modifies the annotated normal of the noisy points faithfully, and thus substantially improves the network robustness against noise without the loss of any shape details.

\begin{figure}[b]
\centering
\includegraphics[scale=0.35]{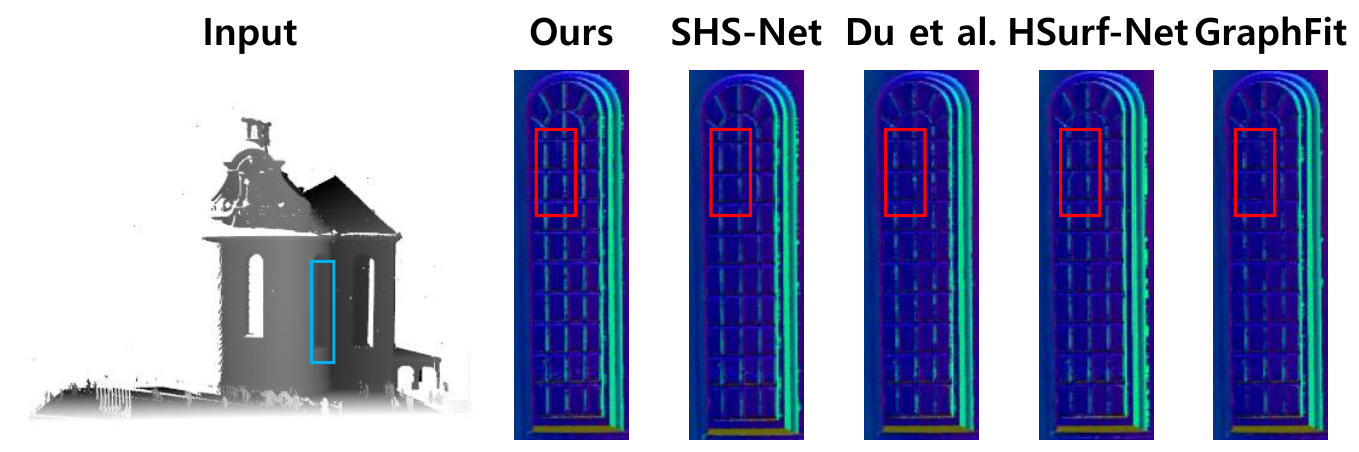}
\vskip -0.3cm
\caption{Qualitative comparisons on the Semantic3D dataset, where point normals are mapped to RGB colors.}
\vskip -0.3cm
\label{fig: Semantic3D}
\end{figure}

\begin{figure}[b]
\centering
\includegraphics[scale=0.27]{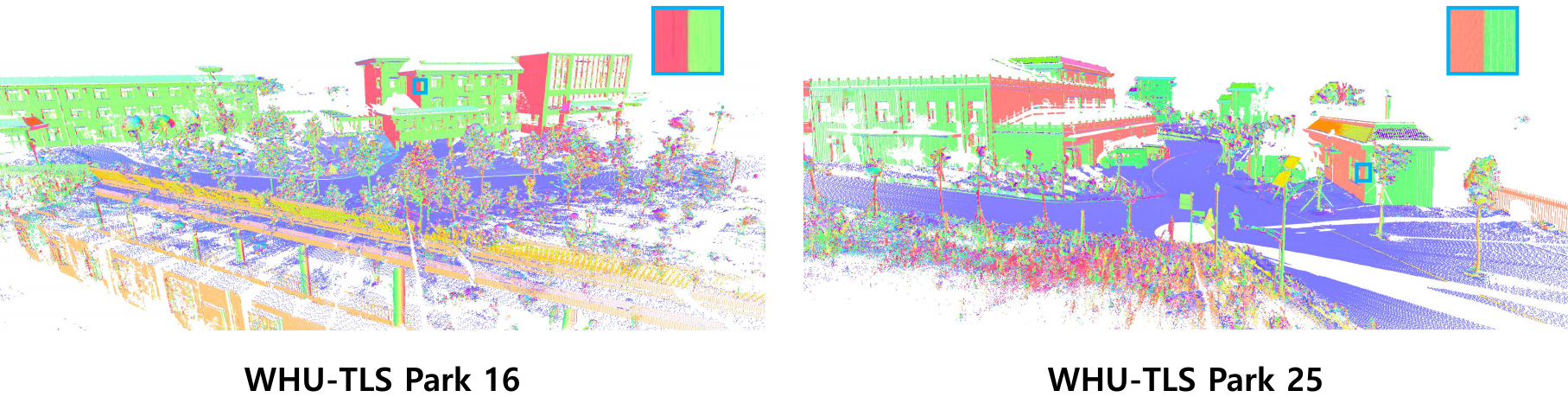}
\vskip -0.3cm
\caption{Qualitative results on the WHU-TLS dataset.}
\vskip -0.3cm
\label{fig: WHU-TLS}
\end{figure}

\subsection{Qualitative Results}

\begin{figure*}[ht]
\centering
\includegraphics[scale=0.38]{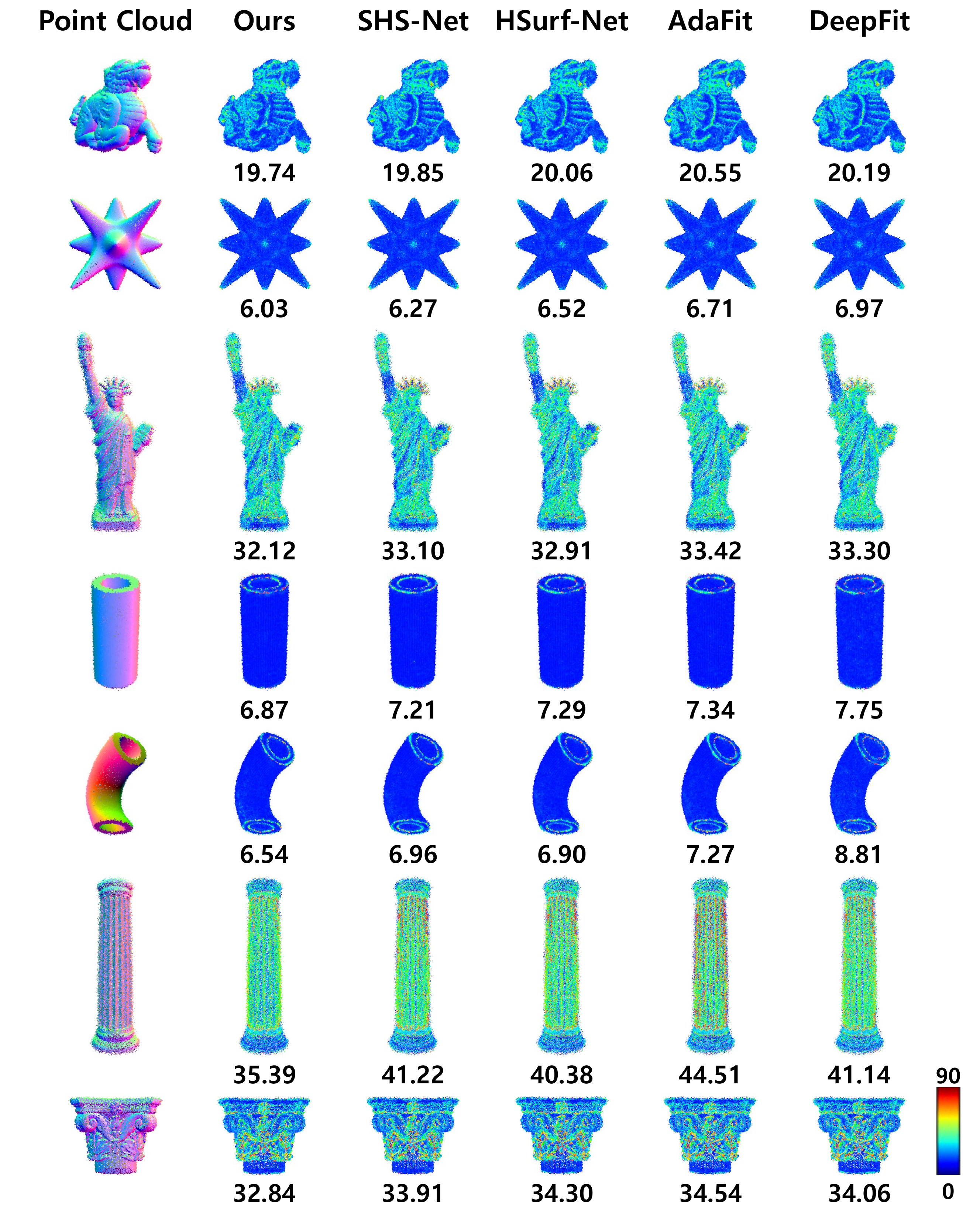}
\caption{Comparisons on the point clouds with heavy noise ($\sigma=0.6\%$) in the PCPNet dataset. We use the heat map to visualize the CND error. Our method achieves the highest accuracy on all models.}
\label{fig: med noise}
\end{figure*}

\begin{figure*}[ht]
\centering
\includegraphics[scale=0.38]{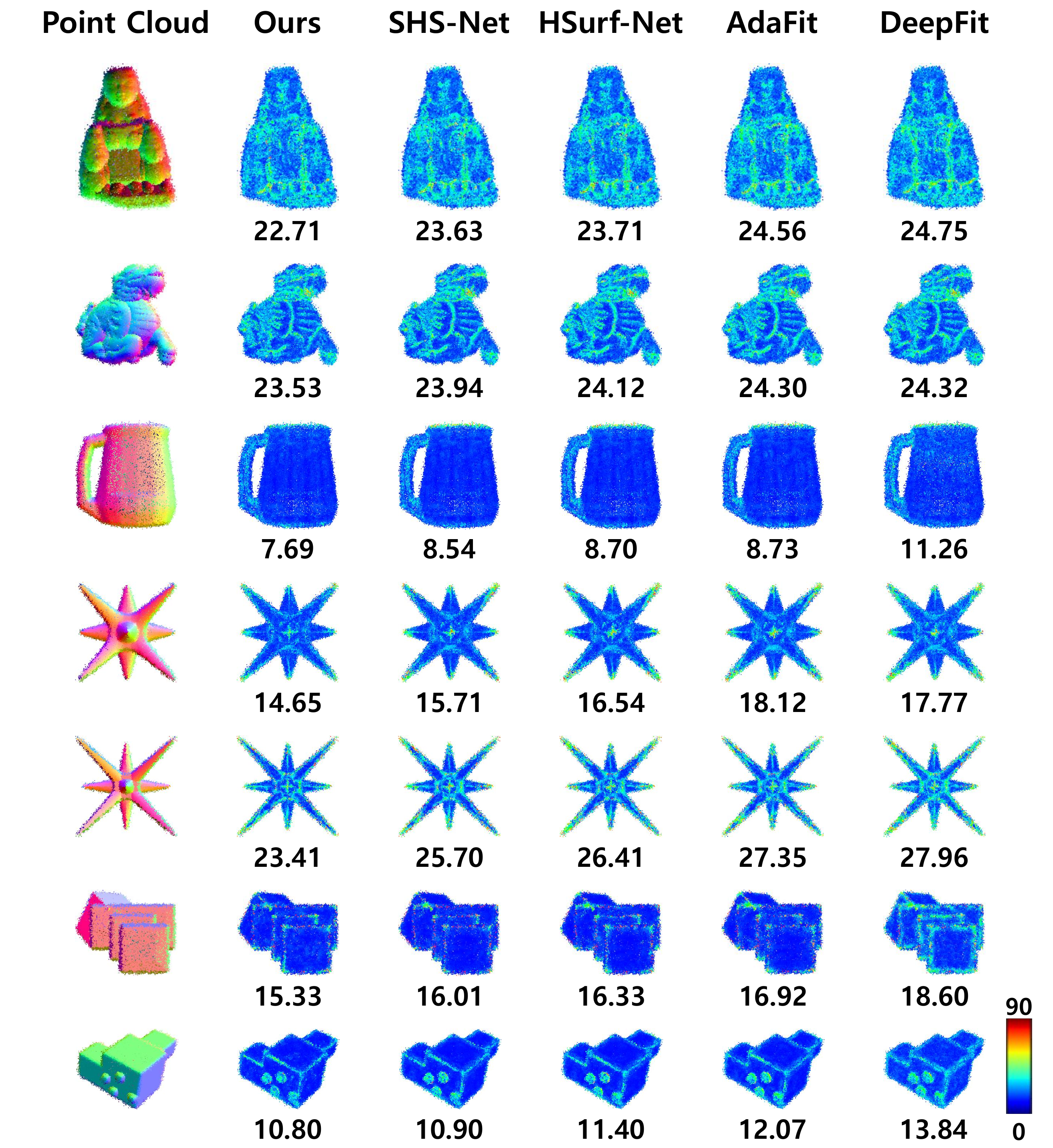}
\caption{Comparisons on the point clouds significantly affected by heavy noise ($\sigma=1.2\%$) in the PCPNet dataset. We use the heat map to visualize the CND error. Our method consistently achieves the highest accuracy across all models.}
\label{fig: high noise}
\end{figure*}

We present more quantitative and qualitative results on the real-word outdoor Semantic3D dataset~\cite{hackel2017isprs} and the LiDAR WHU-TLS dataset~\cite{dong2020registration} in Fig.~\ref{fig: Semantic3D} and Fig.~\ref{fig: WHU-TLS}, and point clouds with heavier noise in Fig.~\ref{fig: med noise} and Fig.~\ref{fig: high noise} to demonstrate the better performance and generalization of our proposed method.

\subsection{Ablation Studies on Real-word Dataset}

As reported in Table~\ref{tab: ablations on scenenn}, we have conducted additional ablation studies on the real indoor dataset SceneNN, to further demonstrate the generalization of each proposed component. The models are trained on the PCPNet dataset and share the same setting with the ones in Sec 5.3. Results regarding each component provide further validation of the technical soundness of our method, suggesting its effectiveness and robustness.

\begin{table}[t]
\centering
\caption{Ablation studies on the realistic dataset SceneNN.}
\renewcommand{\arraystretch}{1.3}
\resizebox{\linewidth}{!}{
\begin{tabular}{cc|cc|c}
\hline
\multicolumn{2}{c|}{Category}                                  & Clean         & Noise          & Averge        \\ \hline
\multirow{3}{*}{(a)} & w/o Local Feature Extration             & 7.58          & 11.18          & 9.38          \\
                     & w/ Single-scale Local Feature Extration & 7.14          & 11.01          & 9.08          \\
                     & w/o Attentional Feature Fusion (AFF)    & 7.06          & 10.86          & 8.96          \\ \hline
\multirow{3}{*}{(b)} & w/o Hierarchical Architecture           & 7.52          & 11.25          & 9.39          \\
                     & w/o Muli-scale Global Feature           & 6.98          & 10.95          & 8.97          \\
                     & w/o Local Feature                       & 6.96          & 11.21          & 9.08          \\ \hline
\multirow{2}{*}{(c)} & w/o Position Feature Fusion (PFF)       & 7.09          & 10.86          & 8.97          \\
                     & w/o Weighted Normal Prediction (WNP)    & 7.06          & 11.09          & 9.07          \\ \hline
\multirow{2}{*}{(d)} & w/o QSTN                                & 7.10          & 11.10          & 9.10          \\
                     & w/o  Z-direction Transformation Loss    & 7.08          & 11.06          & 9.07          \\ \hline
(e)                  & w/o CND                                 & 6.94          & 10.96          & 8.95          \\ \hline
\multicolumn{2}{c|}{ours}                                      & \textbf{6.92} & \textbf{10.82} & \textbf{8.87} \\ \hline
\end{tabular}}
\label{tab: ablations on scenenn}
\end{table}

\begin{table}[t]
\centering
\caption{Timings on realistic indoor dataset SceneNN with 10K points per scene.}
\renewcommand{\arraystretch}{1.3}
\resizebox{\linewidth}{!}{
\begin{tabular}{c|cccc|c}
\hline
Data (10K) & No.032 & No.207 & No.032-Noise & No.207-Noise & Ave. \\ \hline
Time (s)    & 4.8    & 3.69   & 3.8          & 3.76         & 4.01    \\ \hline
\end{tabular}}
\label{tab: time}
\end{table}

\section{More Applications}

In this section, we provide more results of employing our method to downstream tasks. Both quantitative and qualitative results demonstrate that our method outperforms competitors,  in both surface reconstruction~\cite{kazhdan2006poisson} and denoising~\cite{lu2020low} tasks.

\subsection{Real-time Application}

Timings reported in Table~\ref{tab: time} show our acceptable efficiency, which indicate that our method is not yet real-time capable, as stated in Sec. 6.

\subsection{Surface Reconstruction}

\begin{figure*}[ht]
\centering
\includegraphics[scale=0.2]{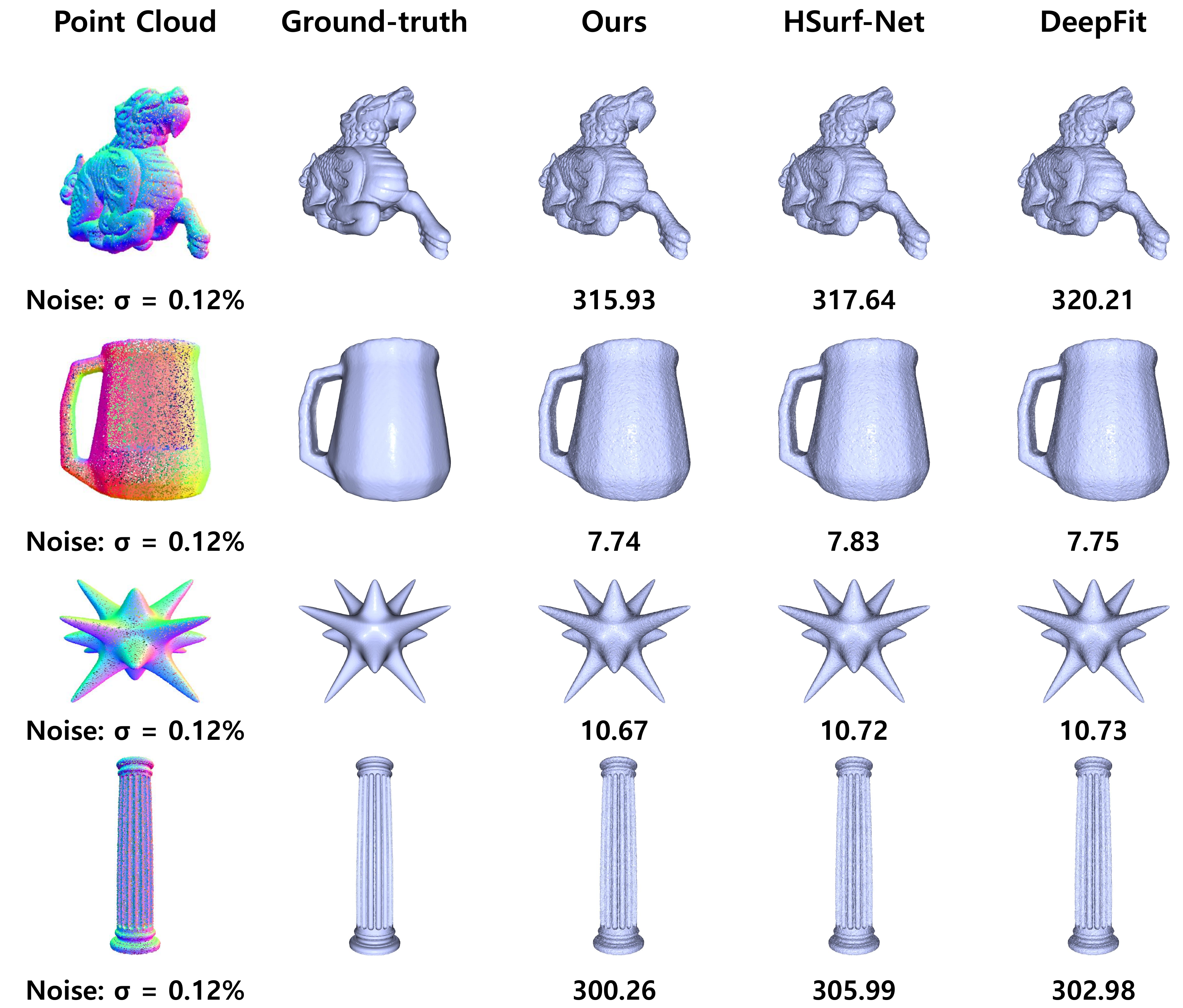}
\caption{Comparisons on the reconstruction results. Our method achieves
the best reconstruction quality.}
\label{fig: more reconstruction}
\end{figure*}

In Fig.~\ref{fig: more reconstruction}, we show more mesh models reconstructed using normals predicted by different methods and the corresponding SMD ($\times10^{-4}$) of reconstructed surfaces. As observed, our method consistently generates accurate reconstruction surfaces on the PCPNet dataset.

\subsection{Denoising}

\begin{figure*}[ht]
\centering
\includegraphics[scale=0.2]{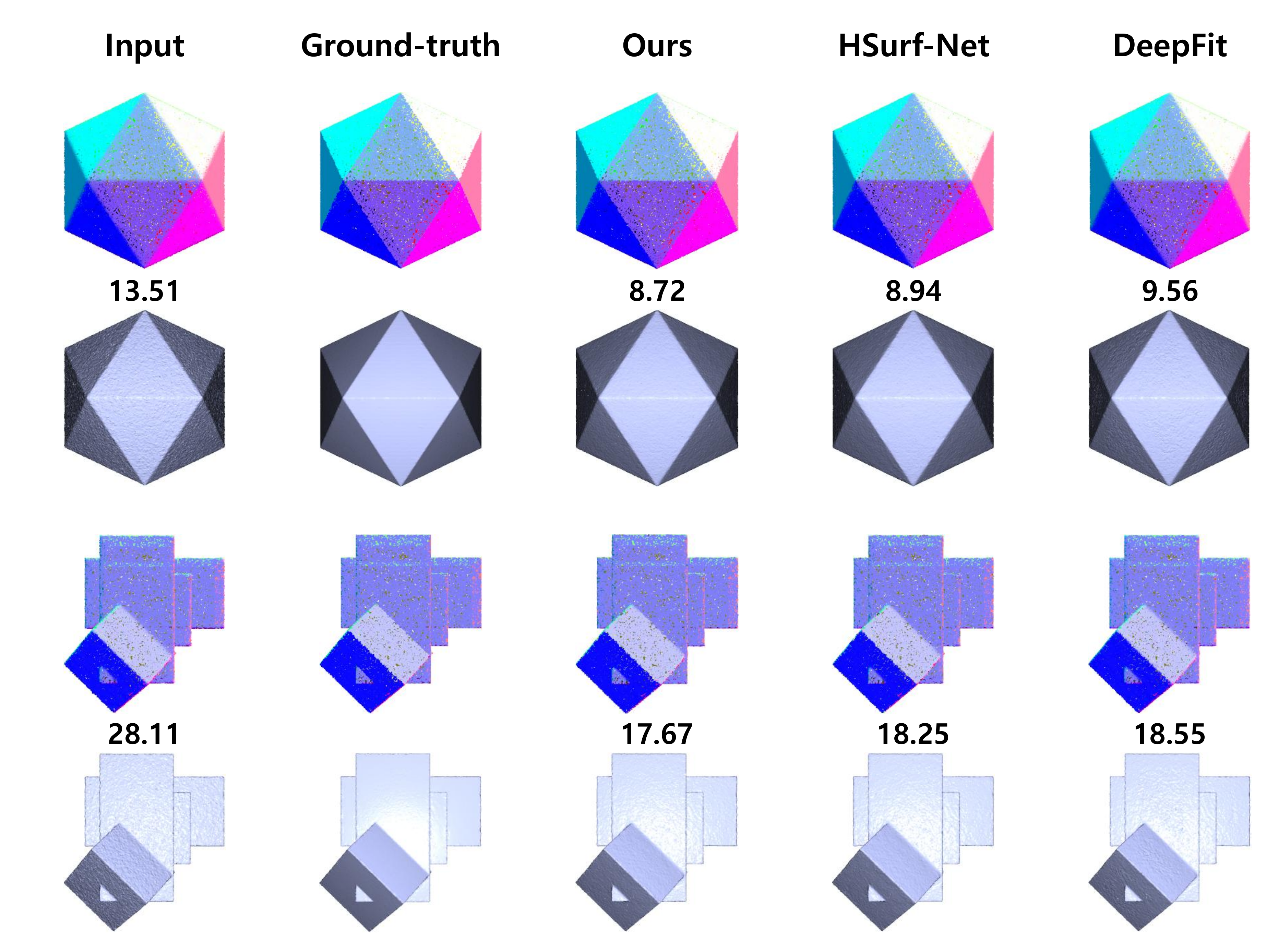}
\caption{Comparisons on the denoising results. The first row shows the denoised point clouds while the second row shows the corresponding reconstructed suraces. Our method achieves the best denoising results along with high-quality reconstruction surfaces.}
\label{fig: more denoising}
\end{figure*}

In Fig.~\ref{fig: more denoising}, we present the denoising results of the instances in the PCPNet dataset using normals estimated by competing approaches, along with the CD ($\times10^{-6}$) and their corresponding reconstructed surfaces. Our method also achieves the best denosing results.

\end{appendices}

\end{document}